\documentclass{article}

\usepackage{arxiv}
\usepackage{makecell}
\usepackage{natbib}
\usepackage{graphicx}
\usepackage{epstopdf, epsfig}
\usepackage{amssymb}
\usepackage{amsmath}
\usepackage{hyperref}
\usepackage[nameinlink, noabbrev]{cleveref}    
\usepackage{xcolor}
\usepackage{subfig}
\usepackage{float}
\usepackage[normalem]{ulem}
\usepackage{soul}
\usepackage{colortbl,xcolor}
\usepackage[ruled,linesnumbered]{algorithm2e}
\usepackage{multirow}
\usepackage{multicol}
\usepackage{tabularx}
\usepackage{setspace}

\hypersetup{
    colorlinks=true,
    linkcolor=blue,
    citecolor=blue,
    filecolor=magenta,      
    urlcolor=cyan,
    hypertexnames=true
    }

\usepackage{booktabs}       
\usepackage{nicefrac}       
\usepackage{microtype}      

\usepackage{ulem}

\title{A Graph-informed Kriging model for sample-efficient Bayesian optimization of wind farm layouts}
\renewcommand{\shorttitle}{GIK model for sample-efficient WFLO}

\author{%
  \parbox{0.9\textwidth}{%
    \centering
    Zhen-Fan Wang \\
    State Key Laboratory of Ocean Engineering, School of Ocean and Civil Engineering, Shanghai Jiao Tong University, Shanghai, 200240, China \\[6pt]
    Kai Zhang\thanks{Corresponding author: kai.zhang@sjtu.edu.cn} \\
    State Key Laboratory of Ocean Engineering, School of Ocean and Civil Engineering, Shanghai Jiao Tong University, Shanghai, 200240, China \\[6pt]
    Xianping Du \\
    School of Ocean Engineering and Technology, Sun Yat-Sen University \& Southern Marine Science and Engineering Guangdong Laboratory (Zhuhai), Zhuhai, 519000, China \\[6pt]
    Zhaolong Han \\
    State Key Laboratory of Ocean Engineering, School of Ocean and Civil Engineering, Shanghai Jiao Tong University, Shanghai, 200240, China \\[6pt]
    Dai Zhou \\
    State Key Laboratory of Ocean Engineering, School of Ocean and Civil Engineering, Shanghai Jiao Tong University, Shanghai, 200240, China \\
    Shenzhen Research Institute of Shanghai Jiao Tong University, Shenzhen, 518063, China \\[6pt]
    Onur Bilgen \\
    Department of Mechanical and Aerospace Engineering, Rutgers University, Piscataway, New Jersey 08854, USA
  }%
}

\begin{document}

\maketitle
\begin{abstract}

Wind farm layout optimization becomes increasingly challenging as the number of turbines grows, particularly when computationally expensive aerodynamic models are used. 
Conventional Kriging-based Bayesian optimization suffers from poor sample efficiency in high-dimensional layout spaces because limited training data provide inadequate coverage of the design domain. 
To address this limitation, we develop a Graph-informed Kriging framework that incorporates reusable aerodynamic interaction information into the surrogate mean function. 
The wind farm is represented as a directed aerodynamic graph, in which pairwise turbine interactions characterized in a low-dimensional relative-coordinate space are assembled into a farm-scale Graph-informed mean. 
This structured mean captures the dominant variation of annual energy production, allowing the Gaussian process to focus on the remaining residual structure and uncertainty. 
Numerical experiments show that Graph-informed Kriging consistently improves annual energy production prediction accuracy over ordinary and generic-trend Kriging models under sparse-data conditions and maintains this advantage as the layout dimension and wake-interaction strength increase. 
When embedded in Bayesian optimization, Graph-informed Kriging substantially improves optimization sample efficiency and achieves competitive solutions for the International Energy Agency Wind Task 37 benchmark using approximately 280 annual energy production evaluations. 
In a controlled 16-turbine benchmark case, the Graph-informed Kriging also outperforms both ordinary and multi-fidelity Kriging when integrated into Bayesian optimization framework under the same high-fidelity evaluation budget. 
Finally, the framework is successfully extended to a substantially larger wind farm with 48 turbines, demonstrating its applicability to large-scale computational-fluid-dynamics-based wind farm layout optimization.
These results demonstrate that Graph-informed surrogate modeling can facilitate the use of higher-fidelity aerodynamic models in large-scale wind farm layout optimization, providing a pathway toward more reliable wind farm layout design.

\end{abstract}

\section{Introduction}
\label{sec:intro}
As the world transitions toward low-carbon energy sources, wind power has become an important component of sustainable electricity generation \citep{iea2021,irena2022,mccoy2024offshore}. A prevailing trend is the deployment of increasingly large turbines and wind farms, which improves economies of scale but also intensifies aerodynamic wake interactions. The reduced wind speed and enhanced turbulence in turbine wakes can substantially decrease downstream power production, leading to wind farm energy losses of approximately $10$--$20\%$ over the project lifetime \citep{barthelmie2007modelling,barthelmie2009modelling}. Wind farm layout optimization (WFLO) therefore seeks to mitigate wake losses and maximize annual energy production (AEP) by optimally positioning turbines while satisfying site and spacing constraints \citep{samorani2014wind,stanley2019massive,reddy2020wind,thomas2023comparison}. For a wind farm with $N$ turbines, the continuous layout is described by $2N$ positional variables, giving rise to a high-dimensional, nonlinear, non-convex, and multimodal optimization problem.

The computational burden of WFLO is determined by both the cost of evaluating the wind farm AEP (the objective function) and the computational demand of the optimization strategy. 
The former depends strongly on the wind farm aerodynamic model employed. Analytical wake models, such as the Jensen top-hat model and Gaussian wake models, describe wake development using prescribed velocity-deficit profiles and empirical or semi-empirical parameters. 
Owing to their low computational cost, these models are widely used to estimate wind farm performance \citep{shakoor2016wake,kaldellis2021critical}.
Although efficient, their simplified wake representations may introduce errors in AEP prediction under complex aerodynamic conditions \citep{shakoor2016wake,ARCHER20181187,WANG2024118508}. 
In contrast, computational fluid dynamics (CFD) provides a more detailed description of wind farm aerodynamics by solving the governing flow equations. 
In recent years, high-fidelity CFD simulations have increasingly been used to investigate wind farm aerodynamics and complex wake interactions \citep{stevens2017flow,Meneveau02012019,shapiro2022turbulence,sharma2024exawind}.
However, the substantially higher cost of each CFD evaluation makes its repeated use within WFLO computationally demanding.
Depending on the wind farm scale, turbulence model, spatial resolution, and computational resources, CFD simulations may require from minutes to days, whereas analytical wake models typically require only seconds.

\begin{table}
\centering
\caption{Representative CFD-based WFLO studies}
\label{tab:cfd_wflo_core}
\scriptsize
\setlength{\tabcolsep}{3.5pt}
\renewcommand{\arraystretch}{1.5}
\begin{tabular}{p{3.0cm} p{2.25cm} p{2.65cm} p{1.55cm}  p{5.5cm}}
\toprule
Study & Flow/turbine model & Optimization strategy & Turbine count  & Main feature  \\
\midrule

\citet{king2017adjoint}
& RANS+ADM
& Discrete-adjoint gradient optimization
& 16
& Adjoint sensitivities are obtained at a cost largely independent of the number of design variables, but the resulting gradient-based optimization remains local and can be sensitive to the initial layout. \\

\citet{antonini2020optimal}
& RANS+ADM
& Continuous-adjoint gradient optimization
& 14
& Extends efficient adjoint-based CFD optimization to complex terrain, while retaining the initialization sensitivity associated with local gradient-based search. \\

\citet{cruz2020wind}
& RANS+ADM
& Genetic algorithm
& 20
& Direct CFD-in-the-loop evolutionary optimization; demonstrates feasibility but also the high evaluation demand of population-based search. \\

\citet{wang2024optimization}
& RANS+ADM
& Kriging-based Bayesian optimization 
& 8 
& Reduces the number of CFD evaluations through Kriging-based BO, but surrogate construction becomes increasingly data-demanding as the layout dimension grows.\\

\citet{bempedelis2024data}
& LES+ADM
& Kriging-based Bayesian optimization
& 16
& Direct LES-in-the-loop BO demonstrates feasibility of high-fidelity surrogate optimization, while also illustrating its substantial sampling cost. \\

\citet{shao2025pce}
& RANS+ADM
& Kriging-based Bayesian optimization 
& 8
& Reduces both the cost of each AEP evaluation and the number of expensive optimization evaluations through PCE and surrogate modeling. \\

\citet{janeippel2026physics}
& LES+ADM
& Physics-informed Kriging Bayesian optimization
& 16
& Standard black-box BO gives only marginal improvement in complex terrain; physics-informed initial sampling is used to obtain competitive layouts within the LES budget. \\

\bottomrule
\end{tabular}
\vspace{1mm}
\begin{minipage}{0.98\textwidth}
{\footnotesize RANS: Reynolds-averaged Navier-Stokes; LES: Large-eddy simulation; ADM: Actuator disk model; PCE: Polynomial chaos expansion}
\end{minipage}
\end{table}

The second major source of computational cost arises from the optimization strategy. Conventional WFLO methods can be broadly classified into gradient-based and gradient-free approaches, both of which have been widely used with computationally inexpensive analytical wake models \citep{mosetti1994optimization,grady2005placement,kusiak2010design,samorani2014wind,shakoor2016wake,stanley2019massive,reddy2020wind,thomas2023comparison,ZHANG2026127895}. Their limitations, however, become more pronounced when coupled with high-fidelity aerodynamic models such as CFD. Representative CFD-based WFLO studies are summarized in Table~\ref{tab:cfd_wflo_core}. Adjoint-based gradient methods are attractive because the cost of obtaining sensitivities with respect to all design variables is largely independent of the number of design variables \citep{king2017adjoint,antonini2020optimal}. However, they remain local optimization methods and can be sensitive to the initial layout in the strongly non-convex and multimodal WFLO landscape. Gradient-free evolutionary methods provide broader global exploration and require no sensitivity information, but typically demand a large number of objective evaluations, making direct CFD-in-the-loop optimization computationally prohibitive \citep{cruz2020wind}.

Surrogate-assisted optimization offers an alternative strategy for reducing the number of expensive objective evaluations while retaining black-box search capability. Among these methods, Bayesian optimization (BO) is particularly suitable for computationally expensive problems. 
BO constructs a probabilistic surrogate, commonly a Kriging or Gaussian process model, from existing observations and iteratively uses an acquisition function to select new evaluations and update the surrogate until a stopping criterion is met \citep{frazier2018tutorial,lei2021bayesian,du2021understanding}.
{When aerodynamic evaluations are inexpensive, as with analytical wake models, reducing the number of objective function evaluations provides only limited computational benefit, and the overhead associated with surrogate construction and updating can become non-negligible. }
In CFD-based WFLO, by contrast, each aerodynamic evaluation is far more expensive than the surrogate update, making BO particularly attractive for reducing the number of high-fidelity simulations \citep{wang2024optimization,bempedelis2024data}, as shown in Table \ref{tab:cfd_wflo_core}.
Recent studies have further introduced polynomial chaos expansion (PCE) and physics-informed initialization to improve computational efficiency of Kriging-assisted WFLO \citep{shao2025pce,janeippel2026physics}.

Existing CFD-based WFLO studies summarized in Table~\ref{tab:cfd_wflo_core} remain relatively modest in scale, and the Kriging-based BO studies considered previously involve no more than 16 turbines.
Scaling Kriging-based BO to larger wind farms remains challenging because a limited training set provides increasingly sparse coverage of the design space as dimensionality grows \citep{loeppky2009choosing}. Maintaining surrogate accuracy therefore generally requires more training samples. This creates two coupled computational burdens. First, the cost of constructing and updating the Kriging model increases as the training set grows \citep{Rasmussen2006Gaussian}. 
Second, each additional training sample requires a new high-fidelity AEP evaluation.
Because AEP is integrated over a discretized set of wind conditions, one CFD-based AEP evaluation requires a separate full-farm flow simulation for each prescribed wind state. 
As a result, increasing the training set simultaneously raises both surrogate-modeling cost and CFD cost. These coupled limitations reduce the scalability of conventional Kriging-based BO and motivate surrogate models that can provide reliable predictions from sparse high-dimensional data.

To address this sample-efficiency challenge, we formulate a Graph-informed Kriging (GIK) framework inspired by the concept of universal Kriging \citep{Rasmussen2006Gaussian,de2020you}. 
In universal Kriging, a non-trivial deterministic mean is introduced to represent major trends, leaving the Gaussian process to model the remaining residual structure. A practical challenge is the construction of an appropriate trend function. Generic polynomial bases are often convenient, but they do not exploit the specific structure of turbine-to-turbine aerodynamic interactions and may become increasingly data-demanding as the input dimension grows.
Multi-fidelity methods such as co-Kriging provide another route by exploiting correlations between low- and high-fidelity models \citep{kennedy2000predicting,forrester2007multi}. In the classical autoregressive formulation, the high-fidelity response is represented through a scaled low-fidelity process together with a Gaussian-process discrepancy correction. This strategy can reduce the demand for high-fidelity data, but its effectiveness depends on the correlation between fidelity levels and may require a substantial number of full-farm low-fidelity evaluations.

\begin{figure}
    \centering
    \includegraphics[width=1.0 \textwidth]{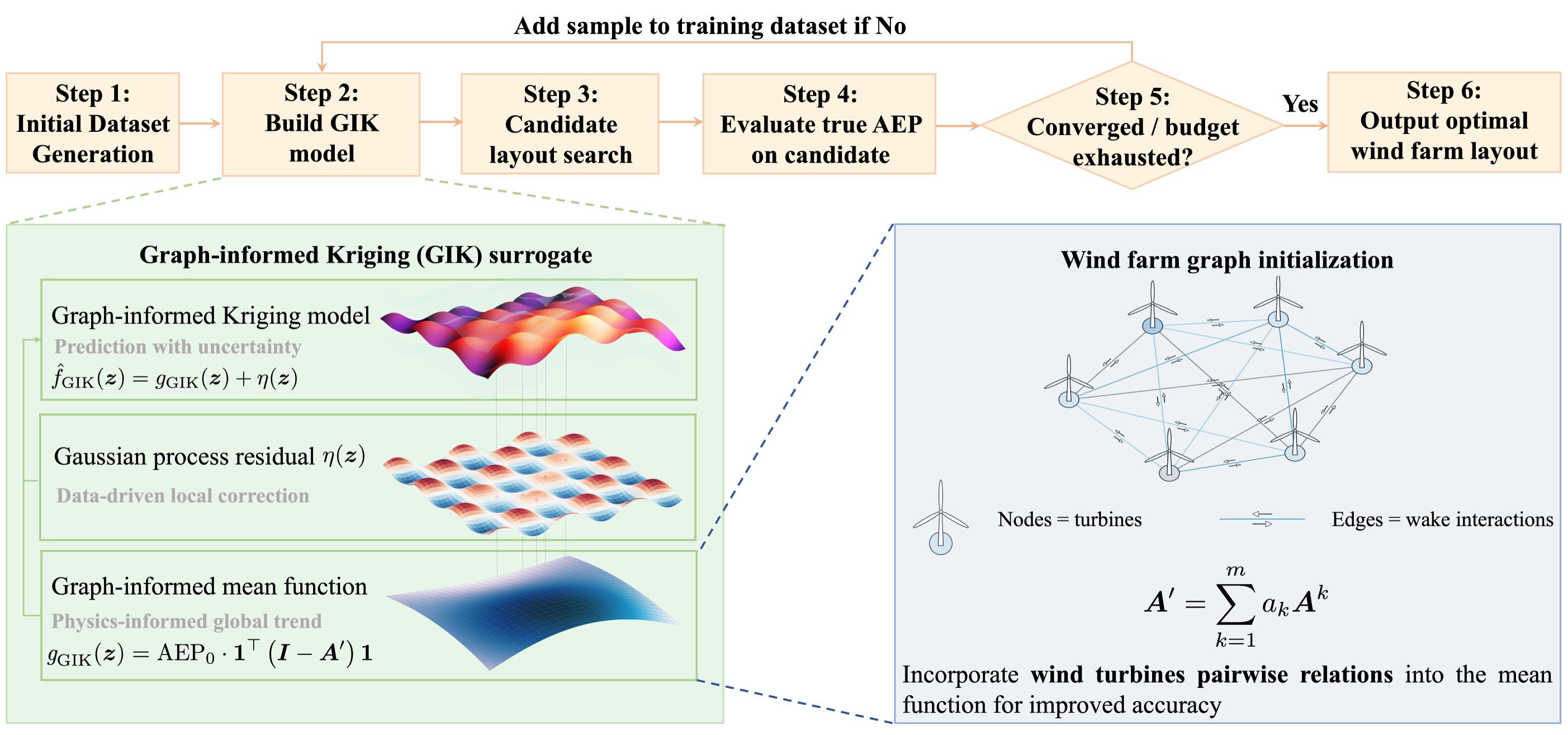}
    \caption{Overview of the GIK-driven Bayesian optimization framework. The framework leverages a graph-theoretic mean function to capture global aerodynamic trends, guiding the efficient exploration of the design space while minimizing AEP evaluations.}
    \label{fig:overview}
\end{figure}

The proposed GIK method addresses these limitations by constructing an aerodynamically informed trend function from reusable pairwise turbine interactions, as illustrated in Figure~\ref{fig:overview}.
The key observation is that WFLO is not an arbitrary high-dimensional black-box problem. Although an $N$-turbine layout is described by $2N$ positional variables, its aerodynamic performance is strongly governed by repeated interactions between turbine pairs. 
For a given wind condition, these interactions can be characterized in a low-dimensional relative-coordinate space, primarily through turbine separation and orientation, and the same interaction relationship can be reused across different turbine pairs and layouts. 
GIK exploits this structure by representing the wind farm as a directed aerodynamic graph and assembling the pairwise interactions into the deterministic mean of the Kriging model. 
The resulting Graph-informed mean provides a structured approximation of the farm-scale AEP variation, while the Gaussian process models the remaining residual structure and uncertainty. 
By transferring reusable low-dimensional aerodynamic information to the farm-scale surrogate, GIK improves prediction accuracy under sparse-data conditions and thereby enhances the sample efficiency of Bayesian optimization without relying on computationally burdensome full-farm high-fidelity simulations.

The remainder of this paper is organized as follows. In \S~\ref{sec:wflo}, the WFLO problem is formulated. The GIK model is introduced in \S~\ref{sec:AEP}, and its predictive performance is then evaluated across different training-set sizes, wind conditions, wind farm densities, and turbine numbers. In \S~\ref{sec:wflo_cases}, the GIK model is incorporated into BO and first examined through parametric studies based on analytical wake models. 
Its optimization performance is subsequently compared with several established algorithms using the IEA Wind Task 37 benchmark. 
The feasibility of GIK-based BO for CFD-based WFLO is further demonstrated using 16- and 48-turbine wind farms. Finally, \S~\ref{sec:conclusion} summarizes the main findings, discusses the implications of the proposed framework, and outlines directions for future research.

\section{Formulation of WFLO problem}
\label{sec:wflo}
The primary objective of wind farm layout optimization (WFLO) is to maximize the annual energy production (AEP) of a wind farm by optimizing the spatial configuration of turbines to mitigate aerodynamic wake losses. 
Mathematically, this optimization problem is formulated as
\begin{equation}
\max_{\boldsymbol{X}} \ \mathrm{AEP}(\boldsymbol{X}; \xi),
\end{equation}
where $\boldsymbol{X}=[\boldsymbol{x},\boldsymbol{y}]^\textrm{T}$ represents the layout vector, with $\boldsymbol{x}=(x_1, \dots, x_N)$ and $\boldsymbol{y}=(y_1, \dots, y_N)$ denoting the coordinates of the turbines. The variable $\xi$ represents the stochastic wind conditions.
Standard practice quantifies AEP by integrating the power output across a discretized probability distribution of wind states (speed and direction). This is calculated as:
\begin{equation}
\label{equ:aep}
\mathrm{AEP}(\boldsymbol{X}; \xi) = T_{\mathrm{yr}} \sum_{k=1}^{N_w} p_k P_{\textrm{farm}}(\boldsymbol{X}, \xi_k),
\end{equation}
where $T_{\mathrm{yr}} \approx 8766$ hours represents the operational time per year (accounting for leap years), $N_w$ is the number of wind states in the wind rose, and $p_k$ is the frequency of occurrence for the $k$-th wind state $\xi_k$. The term $P_{\textrm{farm}}(\boldsymbol{X}, \xi_k) = \sum_{i=1}^N P_i(\boldsymbol{X}, \xi_k)$ denotes the aggregate power output of the farm, where $P_i$ is the power generated by turbine $i$.
To estimate wind farm power, this study uses two distinct modeling approaches tailored to specific validation stages. A low-fidelity analytical Gaussian wake model (Section \ref{sec:gaussian}) is employed to facilitate extensive parametric studies of the proposed GIK model. Subsequently, an actuator-disk-based Reynolds-averaged Navier-Stokes approach (ADM-RANS, detailed in Section \ref{sec:adm}) is utilized to demonstrate the efficacy of the framework in solving realistic, CFD-based WFLO problems.

The optimization is subject to geometric constraints to ensure site compliance and operational safety. Turbines must be located within the designated farm boundary $\mathcal{D}$, and a minimum separation distance $d_{\textrm{min}}$ must be maintained between any pair of turbines to prevent mechanical fatigue and overlapping. These constraints are defined as:
\begin{align}
\text{subject to: } \quad & (x_i, y_i) \in \mathcal{D}, \quad \forall i \in \{1, \dots, N\}, \\
& |(x_i, y_i) - (x_j, y_j)|_2 \geq d_{\textrm{min}}, \quad \forall i \neq j,
\end{align}
where $d_{\textrm{min}}$ is set to $2D$, with $D$ denoting the rotor diameter.

\section{Graph-informed Kriging model for prediction of AEP}
\label{sec:AEP}

This section presents the proposed Graph-informed Kriging (GIK) model, in which pairwise turbine interaction information is incorporated into the Kriging mean through a graph-based wind farm representation. 
Its predictive performance is then evaluated under different wind conditions and wind farm configurations.

\subsection{Conventional Kriging model}
The Kriging model \citep{krige1951statistical} serves as the foundational non-parametric probabilistic model for this study.
It is defined by a linear combination of a deterministic mean function and a stochastic process, formulated as
\begin{equation}
    \hat{f}(\boldsymbol{z}) =\textit{g}(\boldsymbol{z}) + \eta(\boldsymbol{z}),
    \label{equ:KG}
\end{equation}
where $\hat{f}(\boldsymbol{z})$ denotes the Kriging approximation, the variable $\boldsymbol{z}$ is the vector containing all design variables, $g(\boldsymbol{z})$ represents the mean function, and $\eta(\boldsymbol{z})$ is a zero-mean Gaussian process representing the residual component.
The mean function $g(\boldsymbol{z})$ is intended to capture the prior information or global trend of the unknown function \citep{de2020you}, while the stochastic term $\eta(\boldsymbol{z})$ accounts for local fine-tuning by modeling the correlations between training samples.

In the standard application of conventional Kriging, the mean function is usually specified as zero, a constant, or a simple linear function due to a lack of prior information regarding the objective function \citep{deisenroth2019practical}. Consequently, the modeling effort focuses primarily on training the Gaussian process residual function, which is defined by the spatial covariance function:
\begin{equation}
    K_{ij} = \mathrm{Cov}[\eta(\boldsymbol{z}^{(i)}),\eta(\boldsymbol{z}^{(j)})] = \sigma_f^{2}R(\boldsymbol{z}^{(i)},\boldsymbol{z}^{(j)}),
    \label{equ:cov}
\end{equation}
where $K_{ij}$ denotes the covariance between the Gaussian-process residuals at the $i$-th and $j$-th training samples, \(\sigma_f^2\) denotes the process variance of the Gaussian-process residual, and $R$ is the correlation. 
While various correlation functions are available, the Gaussian correlation function is the most widely adopted and is defined as:
\begin{equation}
     R(\boldsymbol{z}^{(i)},\boldsymbol{z}^{(j)}) = \mathrm{exp}\left(-\sum_{k=1}^{2N}\lambda_k(\boldsymbol{z}^{(i)}_{k}-\boldsymbol{z}^{(j)}_{k})^{2}\right), \ \ \ i,j\ =\ 1,2,...,n, 
\end{equation}
where \(\lambda_k>0\) is the inverse squared length-scale parameter associated with the \(k\)-th design variable, determined via maximum likelihood estimation \citep{KrigingTrain}, and $n$ is the sample size.
In this study, the \verb|GPyTorch| Python package \citep{gardner2018gpytorch} is used to train these parameters.

As mentioned previously, the conventional Kriging model suffers from the curse of dimensionality in high-dimensional optimization problems \citep{hou2022dimensionality,song2024improved}. 
As the number of turbines increases, the dimensionality of the layout space grows linearly, while a fixed number of samples provides progressively poorer coverage of that space.
The resulting distance concentration weakens the ability of local covariance kernels to distinguish informative neighborhood structure in sparse high-dimensional data \citep{bengio2005curse}. 
These limitations can substantially reduce the sample efficiency and scalability of conventional Kriging models in large-scale WFLO, motivating the aerodynamically informed surrogate formulation developed in this work.

\subsection{Graph-informed Kriging model}
\label{sec:gikFormulation}
To overcome the limitations of the conventional Kriging model, we introduce a Graph-informed Kriging (GIK) model that integrates explicit physical priors into the mean function, thereby reducing the reliance on large datasets for learning the global landscape of the objective function.
The fundamental insight embedded in the GIK model is that the propagation of aerodynamic wakes exhibits characteristics typical of complex networked systems, where local pairwise interactions cascade into global patterns of power loss, analogous to influence propagation in social, biological, or fluid dynamics \citep{network2010newman,estrada2008communicability,battiston2021physics,TAIRA2022100823}.
By modeling how the relative position of turbine pairs dictates these aerodynamic interactions, we construct a deterministic trend function that resolves the primary wake losses. 
This graph-based mean function provides a robust global baseline, allowing the Kriging model to focus its data-driven learning on residual complexities rather than learning fundamental aerodynamic physics from scratch.

\begin{figure}
    \centering
    \includegraphics[width=0.5 \textwidth]{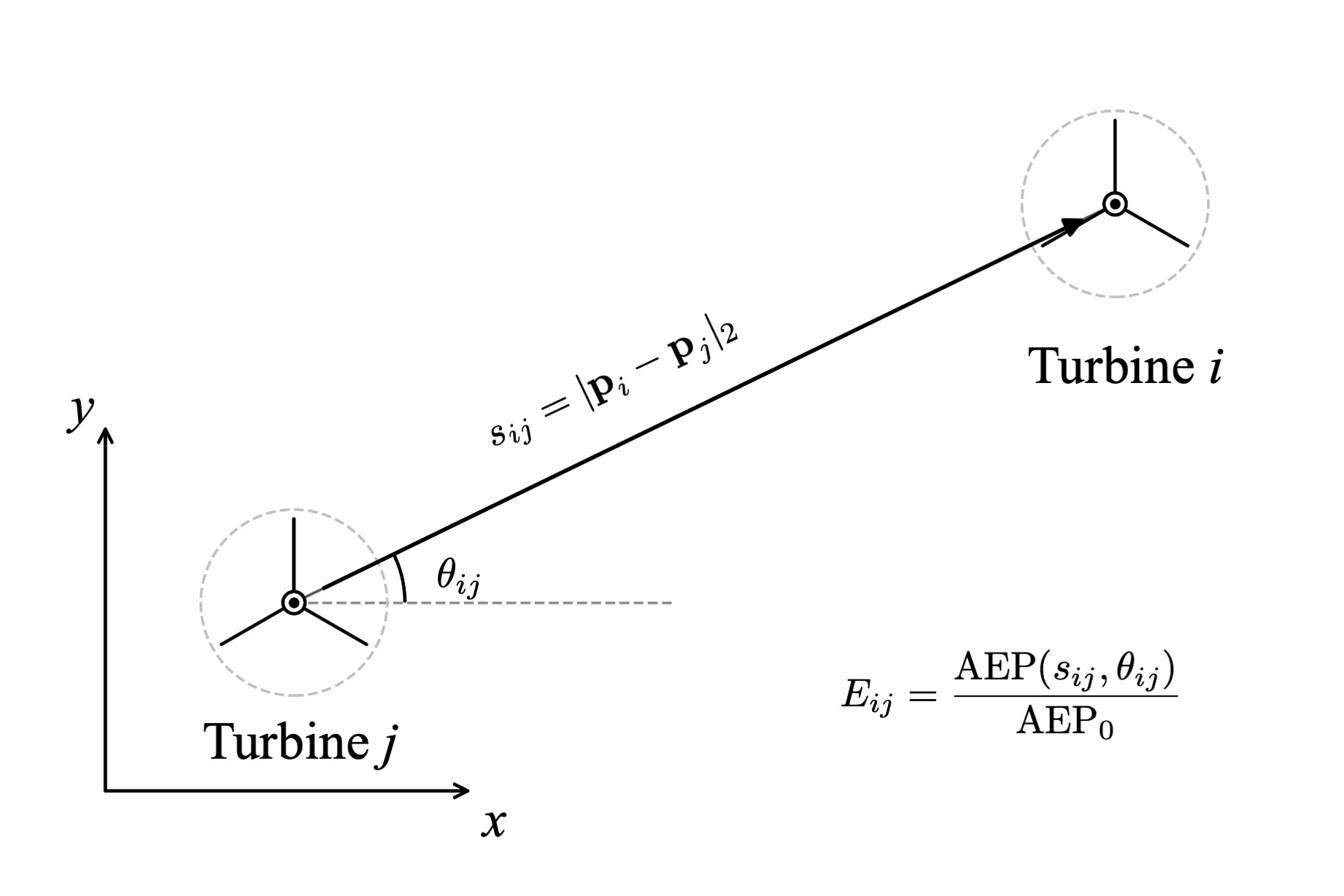}
    \caption{Spatial relations of a turbine pair used in the Graph-informed model.}
    \label{fig:s_theta_illustration}
\end{figure}

To operationalize this idea, we formalize the wind farm as a directed weighted graph $\mathcal{G} = (\mathcal{V}, \mathcal{E})$.
The vertex set $\mathcal{V} = \{\mathbf{p}_1, \dots, \mathbf{p}_N\}$ where  $\mathbf{p}_i=[x_i,y_i]$ encodes the spatial state (coordinates) of each turbine. 
The directed edge set is defined as $\mathcal{E}=\{(\mathbf{p}_j,\mathbf{p}_i)\mid i,j=1,\dots,N,\ i\neq j\}$, where each edge $(\mathbf{p}_j,\mathbf{p}_i)$ represents the pairwise aerodynamic interaction from turbine $j$ to turbine $i$, with the edge weights geometrically induced by the relative spatial configuration of node pairs and the wind rose.
We define a normalized pairwise production matrix $\boldsymbol{E} \in \mathbb{R}^{N \times N}$, where each entry $E_{ij}$ illustrated in Fig. \ref{fig:s_theta_illustration}, quantifies the relative efficiency of turbine $i$ operating in the wake of turbine $j$, computed as
\begin{equation}
E_{ij} = \frac{\mathrm{AEP}(s_{ij}, \theta_{ij})}{\mathrm{AEP}_0},
\end{equation}
where $s_{ij} = |\mathbf{p}_i - \mathbf{p}_j|_2$ denotes the separation distance, $\theta_{ij}$ represents the azimuthal angle of the pairwise connection vector, and $\mathrm{AEP}_0$ is the annual energy production of an individual turbine without wake effects.
Finally, the aerodynamic adjacency matrix $\boldsymbol{A}$, which represents the direct wake penalty, is formulated as
\begin{equation}\boldsymbol{A} = \boldsymbol{J}_N - \boldsymbol{E},
\label{equ:A_matrix}
\end{equation}
where $\boldsymbol{J}_N$ is an all-ones matrix.

Since the aerodynamic adjacency matrix \(\boldsymbol{A}\) contains up to \(N(N-1)\) directed pairwise interactions, evaluating each entry through a separate aerodynamic calculation would become increasingly expensive as the number of turbines increases, particularly when more expensive high-fidelity models are employed. To circumvent this, we first construct an offline database that characterizes the aerodynamic interaction between a pair of turbines as a function of their separation distance \(s\) and relative azimuthal angle \(\theta\). For the analytical-wake-model cases considered in this study, the pairwise interaction data are generated using the Gaussian wake model, whereas for the CFD-based cases they are obtained from two-turbine ADM-RANS simulations. In both cases, the resulting discrete database is generated only once and is subsequently used to construct a low-dimensional pairwise-interaction surrogate.

The surrogate exploits the smooth spatial variation of wake-induced turbine interactions. The dependence on the separation distance \(s\) is represented by discretizing the radial coordinate and obtaining the radial weight vector \(\boldsymbol{c}(s)\) through linear interpolation, while the azimuthal dependence is represented using a truncated Fourier basis \(\boldsymbol{\phi}(\theta)\). The pairwise interaction is therefore approximated as
\begin{equation}
A(s,\theta)=\boldsymbol{\phi}(\theta)^{\top}\boldsymbol{c}(s). 
\label{equ:nestedSurrogate} \end{equation}
The surrogate is trained using pairwise data sampled at radial and azimuthal increments of \(1D\) and \(10^\circ\), respectively, with interactions beyond \(15D\) neglected \citep{shaler2019benchmarking,farrell2021design}. Once constructed, it provides rapid algebraic estimates of \(A(s_{ij},\theta_{ij})\) for arbitrary turbine pairs and is used to assemble the adjacency matrix throughout the layout optimization. 

Before analyzing network propagation, we must regularize the graph topology to handle spatial singularities. 
When two nodes become spatially coincident (i.e., $s_{ij} \to 0$), the resulting redundancy (self-loops or duplicate nodes) can distort the flow of influence. 
We introduce a projection operator to decouple these redundant vertices. 
By identifying the set of redundant indices $\mathcal{R} = \{j \mid \exists i < j, s_{ij} = 0\}$, we define $\boldsymbol{P} \in \mathbb{R}^{N \times N}$ as a diagonal matrix that nullifies connections to redundant nodes:
\begin{equation}
    P_{kk} = 
    \begin{cases} 
        0, & \text{if } k \in \mathcal{R}, \\
        1, & \text{otherwise}.
    \end{cases}
\end{equation}
The adjacency matrix $\boldsymbol{A}$ is then regularized via the transformation:
\begin{equation}
\boldsymbol{A} \leftarrow \boldsymbol{P} \boldsymbol{A} \boldsymbol{P} + (\boldsymbol{I} - \boldsymbol{P}),
\end{equation}
where $\boldsymbol{I}$ is the identity matrix.

The adjacency matrix $\boldsymbol{A}$ describes only the direct interaction between each pair of turbines. In a wind farm, however, a downstream turbine may also be influenced indirectly by turbines further upstream through the modified inflow of intermediate turbines. Inspired by the concept of network communicability \citep{estrada2008communicability,network2010newman}, we account for such indirect interactions by including powers of $\boldsymbol{A}$:
\begin{equation}
\label{equ:correction}
\boldsymbol{A}' = \sum_{k=1}^{m} a_k\boldsymbol{A}^k,
\end{equation}
where $\boldsymbol{A}^k$ aggregates all weighted walks of length $k$ between turbine nodes. The first-order term represents the direct pairwise interactions, while the higher-order terms serve as compact graph-based descriptors of how these interactions combine at the farm scale, rather than as explicit models of multi-turbine wake physics. The coefficients $a_k$ are estimated from the full-farm training data by least-squares regression.

Finally, to estimate the energy yield of the wind farm, we compute the weighted in-degree centrality for each node within the global communicability graph. 
The total aerodynamic penalty for node $i$ is the sum of all incoming aggregated influences $A'_{ij}$. 
The Graph-informed Kriging mean function, $\textit{g}_\mathrm{GIK}(\boldsymbol{z})$, is thus formulated as 
\begin{equation}
\textit{g}_\mathrm{GIK}(\boldsymbol{z}) = \mathrm{AEP}_0\cdot \boldsymbol{1}^{\top}\left(\boldsymbol{I}-\boldsymbol{A}'\right )\boldsymbol{1}.
\label{equ:gikmean}
\end{equation}
This formulation essentially subtracts the total cumulative wake penalty from the theoretical maximum capacity of the unconnected network.
The Graph-informed mean is invariant to arbitrary permutations of turbine indices, since turbine relabeling only permutes the rows and columns of the graph matrices without changing the final scalar aggregation in Eq. \ref{equ:gikmean}.

It is important to clarify that the Graph-informed mean function serves as an aerodynamically informed topological prior, designed to capture the global trend rather than to provide a standalone high-accuracy surrogate.
Embedding this approximate landscape effectively detrends the surrogate-based optimization problem, which allows the Kriging to devote more of its modeling capacity to the remaining residual structure and quantify uncertainty. 
This approach is distinct from the recent Graph Neural Network (GNN) studies \citep{PARK2019115883,Li2024GNN}, which aim for high predictive accuracy through complex feature learning but typically demand extensive training datasets. 
In contrast, the GIK framework leverages physical priors to ensure efficient wind farm layout optimization even with sparse data.

\subsection{Performance of $\textit{g}_\mathrm{GIK}(\boldsymbol{z})$ in predicting AEP}
\begin{figure}
    \centering
    \includegraphics[width=0.7\textwidth]{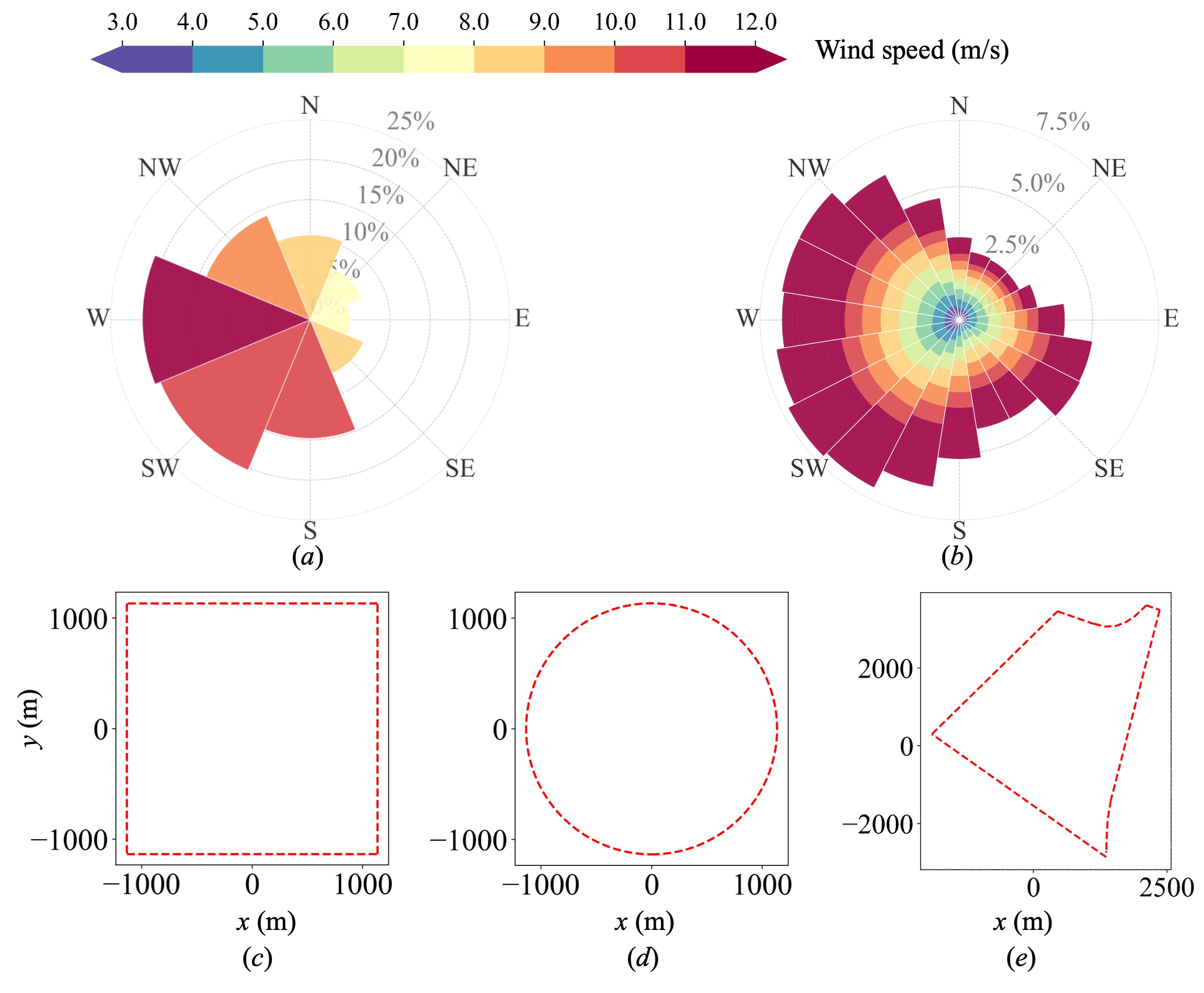}
    \caption{The wind rose and wind farm boundary for case studies: $(a)$ Wind 1, characterized by a unimodal directional distribution; $(b)$ Wind 2, characterized by a multidirectional distribution; $(c)$ square wind farm boundary (Boundary 1); $(d)$ circular wind farm boundary (Boundary 2); and $(e)$ irregular non-convex wind farm boundary (Boundary 3).}
    \label{fig:LFsettings}
\end{figure}

Prior to integrating the mean function $\textit{g}_\mathrm{GIK}(\boldsymbol{z})$ into the full GIK model in equation (\ref{equ:KG}), we verify whether this model is capable of accurately capturing the global trend of wind farms' AEP.
To this end, this study performs comprehensive tests across two distinct wind roses (shown in Figure \ref{fig:LFsettings} ($a$) and ($b$)) within a $30D\times30D$ rectangular area.
For simplicity, the wind farm area is split into $1D\times 1D$ square cells as candidate positions for wind turbines, formulating a discrete problem.
The NREL 5 MW reference wind turbine \citep{jonkman2009definition} is used for the numerical experiments in this study.
For each scenario, the number of turbines $N$ is set to 50, and 20$N$ random wind farm layouts are generated for model training and validation.
The Gaussian wake model is employed to obtain the reference AEPs of wind farms defined by these sampled layouts.
The dataset is randomly divided into two equal-sized subsets as training and testing sets. 
To ensure statistical robustness and mitigate the influence of random splitting bias, this procedure is repeated independently for 10 trials.

After data for training and test is prepared, the $\textit{g}_\mathrm{GIK}(\boldsymbol{z})$ model uses the mean squared error (MSE) as loss function, formulated as
\begin{equation}
    \mathcal{L}_{\text{MSE}} = \frac{1}{n_{\mathrm{tr}}} \sum_{i=1}^{n_{\mathrm{tr}}} \left( f_i - \hat{f}_i(a_1,a_2,...,a_m) \right)^2, 
\label{equ:loss_mse}
\end{equation}
where $f_i$ is the observed AEP output for the $i$-th sample and $\hat{f}_i$ is the corresponding predicted value from the Graph-informed mean function (equation \ref{equ:gikmean}), and $n_{\mathrm{tr}}$ is the training sample size.
Since $g_{\mathrm{GIK}}(\boldsymbol{z})$ is linear with respect to $a_k$, the coefficients are determined efficiently through least-squares regression.
Model performance is evaluated using two standard metrics: the Mean Absolute Percentage Error (MAPE), which measures the relative prediction accuracy, and the Coefficient of Determination ($R^2$), which evaluates the model's capability to capture trend variations.
These are defined as follows:
\begin{subequations}
\begin{align}
    \text{MAPE} &= \frac{100\%}{n_\mathrm{te}} \sum_{i=1}^{n_\mathrm{te}} \left| \frac{f_i - \hat{f}_i}{f_i} \right|,\\
    R^2 &= 1 - \frac{\sum_{i=1}^{n_\mathrm{te}} (f_i - \hat{f}_i)^2}{\sum_{i=1}^{n_\mathrm{te}} (f_i - \bar{f})^2},
\end{align}
\end{subequations}
where $n_\mathrm{te}$ is the size of the test sample, and \(\bar f\) denotes the mean of the reference AEP values over the test set. To verify whether the higher-order terms in equation (\ref{equ:correction}) effectively enhance model performance, we conduct a sensitivity analysis with coupling orders $m \in \{1, 2, 3, 5\}$.

\begin{figure}
    \centering
    \includegraphics[width=1.0 \linewidth]{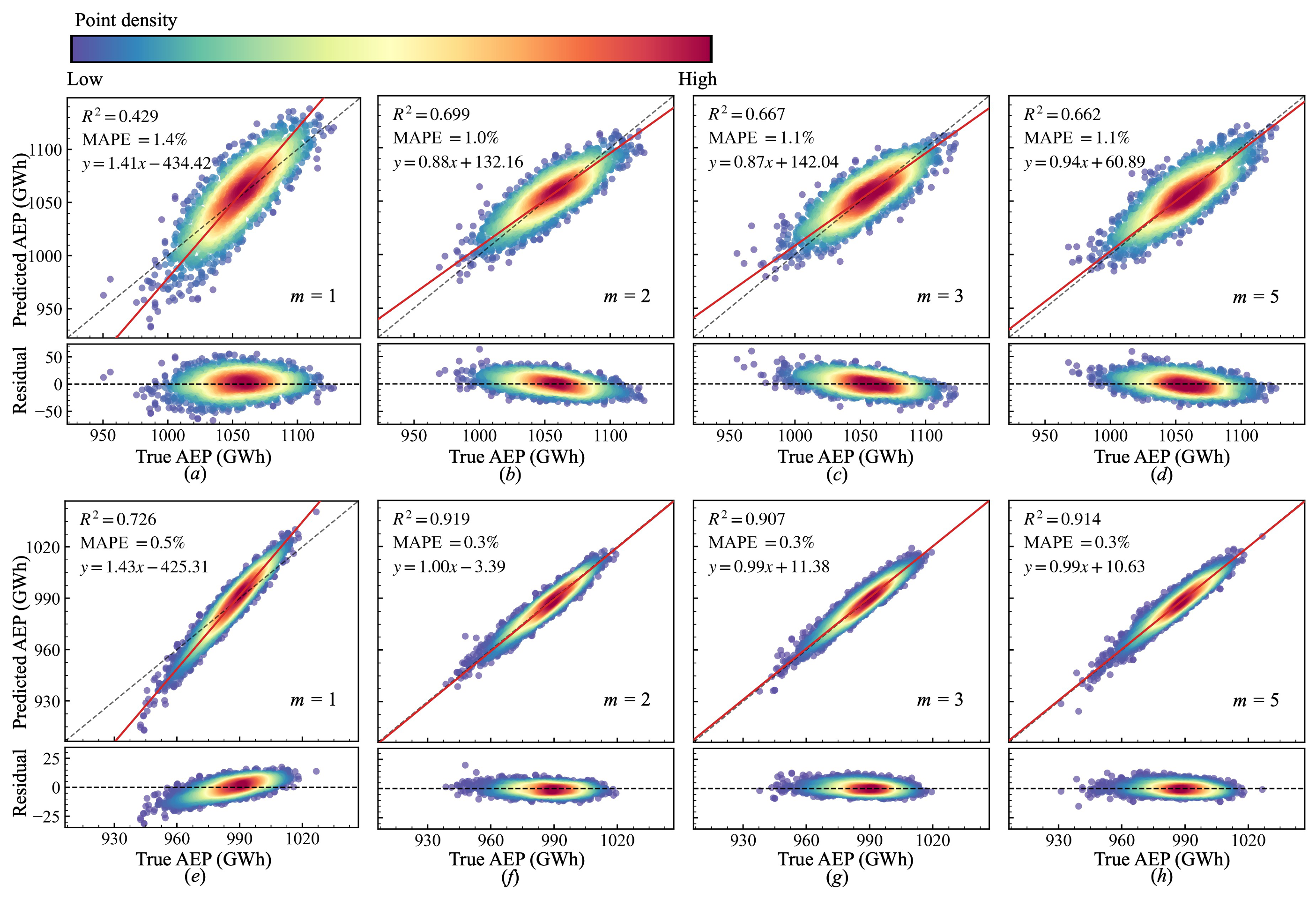}
    \caption{Predictive performance of the $g_{\mathrm{GIK}}(\boldsymbol{z})$ model under varying coupling orders $m$. Top row ($a-d$) and bottom row ($e-h$) correspond to wind scenarios 1 and 2, respectively. Scatter plots show predicted versus true AEP, where the dashed diagonal line denotes perfect agreement. The solid red line indicates the linear fit obtained using Orthogonal Distance Regression. Color intensity represents point density estimated via a Gaussian kernel. }
    \label{fig:R2_onlyTrend}
\end{figure}

Figure~\ref{fig:R2_onlyTrend} compares the performance of $g_\mathrm{GIK}(\boldsymbol{z})$ with different coupling orders $m$ in estimating the wind farm AEP.
With $m=1$, $g_\mathrm{GIK}(\boldsymbol{z})$ yields $R^2$ values of 0.429 and 0.726 (MAPEs of $1.4\%$ and $0.5\%$) for the two wind scenarios, respectively.
In this case, the AEP is estimated as a linear superposition of the direct pairwise interactions in $\boldsymbol{A}$, which may not adequately account for the combined influence of multiple upstream turbines.
Including the second-order term ($m=2$) noticeably improves the prediction, with $R^2$ increasing to 0.699 and 0.919 and MAPE decreasing to $1.0\%$ and $0.3\%$.
Further increasing the coupling order brings no noticeable improvement, possibly because all graph terms are constructed from the same matrix $\boldsymbol{A}$ and the higher-order terms provide limited additional information.
Therefore, coupling orders higher than $m=2$ are not considered in the subsequent analyses.
It is reminded here that $g_\mathrm{GIK}(\boldsymbol{z})$ is not intended to predict AEP accurately on its own.
Rather, it serves to capture the overall AEP trend at low cost, while the Gaussian-process residual in the full GIK model accounts for the remaining discrepancies, as examined in the next section.

\subsection{Comparison with conventional Kriging models}
In this section, we incorporate the Graph-informed mean function $\textit{g}_\mathrm{GIK}(\boldsymbol{z})$ into the full Kriging model, and compare its performance with ordinary Kriging employing a constant mean and universal Kriging models employing linear and polynomial trend functions.
The four variants of Kriging use the same covariance kernel and hyperparameter optimization procedure, such that the principal difference among them lies in the specification of the mean function.
They are implemented and trained using the negative log-marginal likelihood as loss function, formulated as
\begin{equation}
    \mathcal{L}_{\mathrm{NLL}}(\boldsymbol{\lambda},\sigma_f^2)
    =
    \frac{1}{2}
    (\boldsymbol{f}-\boldsymbol{g})^\top
    \boldsymbol{K}^{-1}
    (\boldsymbol{f}-\boldsymbol{g})
    +
    \frac{1}{2}\log|\boldsymbol{K}|
    +
    \frac{n}{2}\log(2\pi),
\label{equ:loss}
\end{equation}
where \(\boldsymbol{K}\in\mathbb{R}^{n\times n}\) is the covariance matrix whose elements are determined from equation (\ref{equ:cov}), \(\boldsymbol{f}=[f^{(1)},\ldots,f^{(n)}]^\top\) is the vector of observed responses, and \(\boldsymbol{g}=[g(\boldsymbol{z}^{(1)}),\ldots,g(\boldsymbol{z}^{(n)})]^\top\) contains the deterministic mean-function values evaluated at the training samples. Here, \(n\) is the number of training samples, and \(\boldsymbol{\lambda}=[\lambda_1,\ldots,\lambda_{2N}]^\top\) denotes the kernel hyperparameters.
For the GIK model, \(g(\boldsymbol{z})=g_{\mathrm{GIK}}(\boldsymbol{z})\), whereas the baseline Kriging models employ their corresponding constant, linear, or polynomial mean functions.
For the training tasks, the Adaptive moment estimation (Adam) optimizer \citep{kingma2014adam} is employed with a learning rate of $0.01$.

\begin{figure}
    \centering
    \includegraphics[width=1.0 \linewidth]{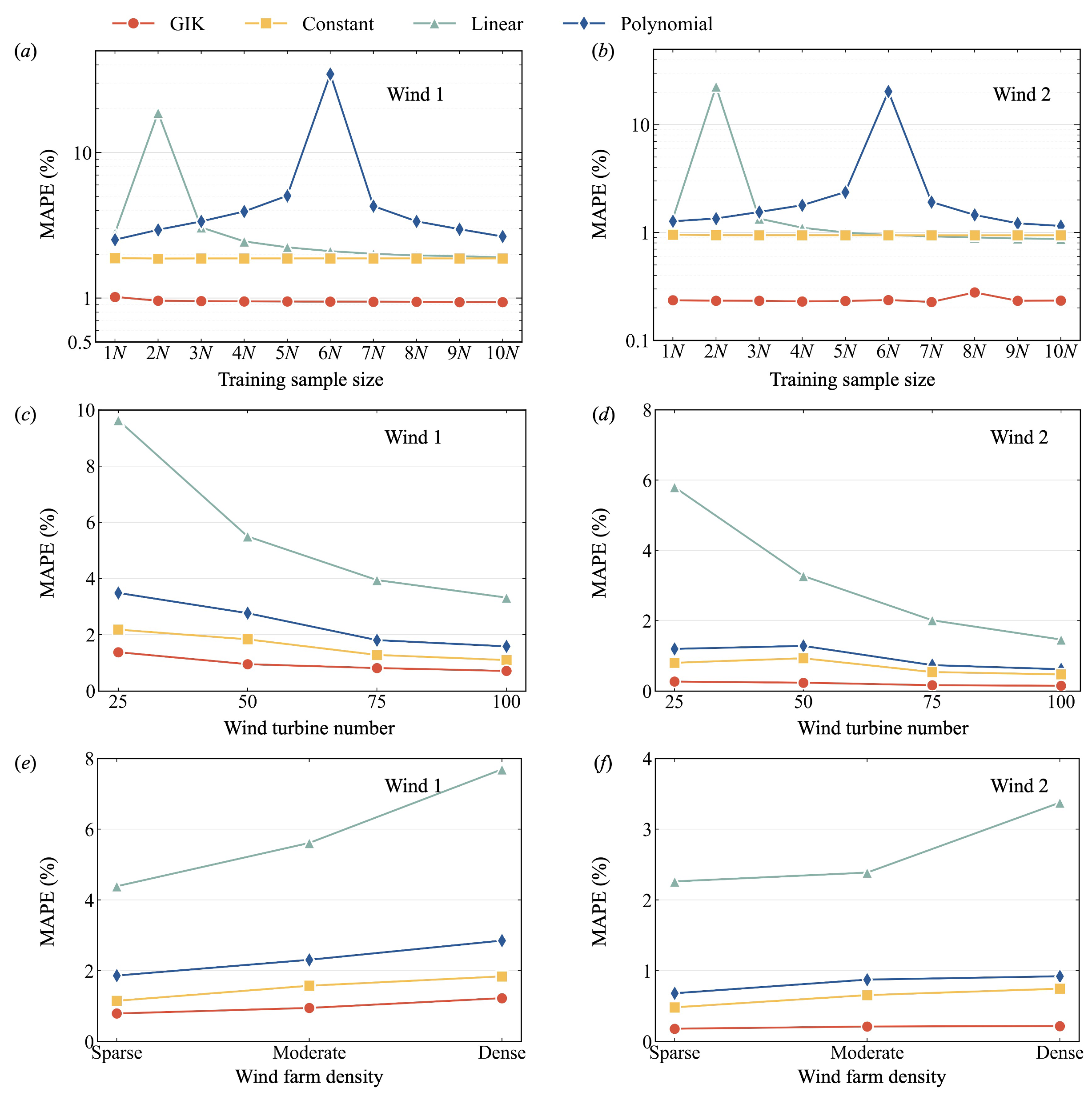}
    \caption{MAPE comparison of GIK and conventional Kriging with constant, linear, and polynomial mean functions for Wind 1 and Wind 2 shown in Figs. \ref{fig:LFsettings}($a$) and ($b$). $(a,b)$ Effect of training sample size from $1N$ to $10N$ with $N=50$. $(c,d)$ Effect of turbine number from 25 to 100 at fixed wind farm density. $(e,f)$ Effect of wind farm density with $N=50$, where the sparse, moderate, and dense cases correspond to $50D\times50D$, $30D\times30D$, and $20D\times20D$ domains, respectively.}
    \label{fig:GIKvsOK}
\end{figure}

We begin the discussion by examining the influence of training sample size on model performance.
Figures \ref{fig:GIKvsOK}($a$-$b$) illustrate the comparative performance of these models in estimating wind farm AEP.
GIK consistently achieves the lowest MAPE across all training-set sizes and both wind scenarios, with only minor variations as the sample size increases from $1N$ to $10N$.
In contrast, the errors of the Kriging baselines remain approximately 2–5 times higher than those of GIK and, for the linear and polynomial trends, exhibit pronounced fluctuations with sample size. 
The corresponding $R^2$ values further support this observation: the Kriging baselines consistently yield negative $R^2$, whereas GIK maintains $R^2>0.727$ and $R^2>0.936$ for Wind 1 and Wind 2, respectively, once the training-set size exceeds $2N$. 
These results indicate that GIK can achieve accurate AEP predictions even when only limited training data are available.

The scalability and robustness of the models are further examined in Figures \ref{fig:GIKvsOK}($c$–$d$), in which the number of turbines is increased from 25 to 100 at a fixed wind farm density, with the training sample size set to $2N$. 
As each turbine contributes two positional variables, the layout-space dimension increases from 50 to 200. 
Despite this increase in dimensionality, GIK consistently yields the lowest MAPE across both wind scenarios. 
In Figure \ref{fig:GIKvsOK} ($e$-$f$), the turbine number is fixed at $N=50$, while the farm domain is reduced from $50D\times50D$ to $20D\times20D$, progressively strengthening wake interactions without changing the problem dimension. 
GIK again maintains the lowest MAPE across all wind farm density levels. 
These results indicate that GIK retains its predictive advantage under both increasing dimensionality and stronger turbine interactions, demonstrating favorable scalability and robustness across a broad range of wind farm setups.

\section{Solving WFLO problems using the BO-GIK framework}
\label{sec:wflo_cases}
We now employ the GIK model to develop a Bayesian optimization (BO) framework (detailed in Appendix \ref{sec:BO}) to evaluate its efficacy for solving the WFLO problems.
In \S~\ref{sec:wflo_lf}, the robustness of BO-GIK is validated against the ordinary-Kriging-based BO framework.
Subsequently, \S~\ref{sec:CompareDirect} benchmarks BO-GIK against established direct optimization methods to assess its competitiveness.
Finally, \S~\ref{sec:wflo_hf} further incorporates the CFD simulations into the BO framework to enable high-fidelity WFLO.

\subsection{Comparison with ordinary Kriging model}
\label{sec:wflo_lf}

We evaluate the performance of our GIK-based framework (BO-GIK) across six WFLO scenarios representing all full-factorial combinations of two wind roses and three boundary geometries (Figure \ref{fig:LFsettings}). 
The wind regimes include a unimodal directional profile (Wind 1, Figure \ref{fig:LFsettings}$a$) and a multi-directional profile (Wind 2, Figure \ref{fig:LFsettings}$b$). 
The spatial domain spans three distinct geometries: a square (Boundary 1, Figure \ref{fig:LFsettings}$c$), a circle (Boundary 2, Figure \ref{fig:LFsettings}$d$), and an irregular non-convex polygon (Boundary 3, Figure \ref{fig:LFsettings}$e$).
To establish a rigorous baseline, we benchmark the proposed approach against Bayesian optimization with ordinary Kriging (BO-OK). 
Given that the Graph-informed mean function $g_\mathrm{GIK}(\boldsymbol{z})$ already provides a substantial improvement in approximation accuracy over OK even when only first-order term is considered ($m=1$), we evaluate both BO-GIK ($m=1$) and BO-GIK ($m=2$) to further assess the contribution of higher-order effects to the optimization performance.
The initial design space consists of $1N$ samples generated using Latin Hypercube Sampling (LHS), and the optimization process is terminated at a maximum of 200 iterations across all cases. 
Each setup is executed across 10 independent runs to evaluate statistical consistency and minimize initialization bias.

\begin{figure}
    \centering
    \includegraphics[width=1.0 \textwidth]{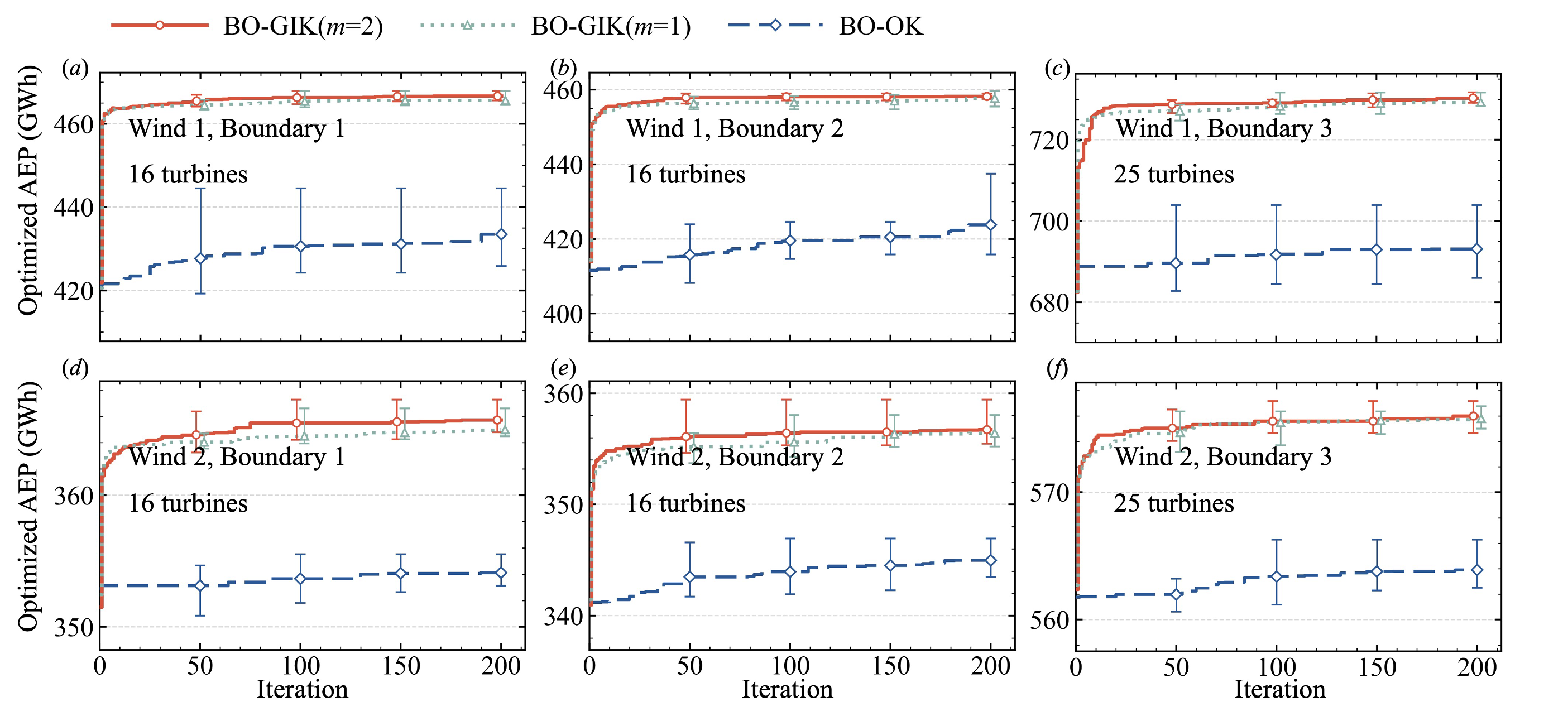}
    \caption{Evolutions of optimized AEPs obtained from BO-GIK and BO-OK. $(a)$--$(c)$ show the results for Wind 1 with Boundaries 1–3, respectively, while $(d)$--$(f)$ show those for Wind 2 with Boundaries 1–3, respectively. The lines represent the mean best-so-far AEP over 10 independent runs, and the bars shown at 50-iteration intervals indicate the corresponding minimum-to-maximum range across the 10 runs.}
    \label{fig:Iteration}
\end{figure}

\begin{figure}
    \centering
    \includegraphics[width=1.0 \textwidth]{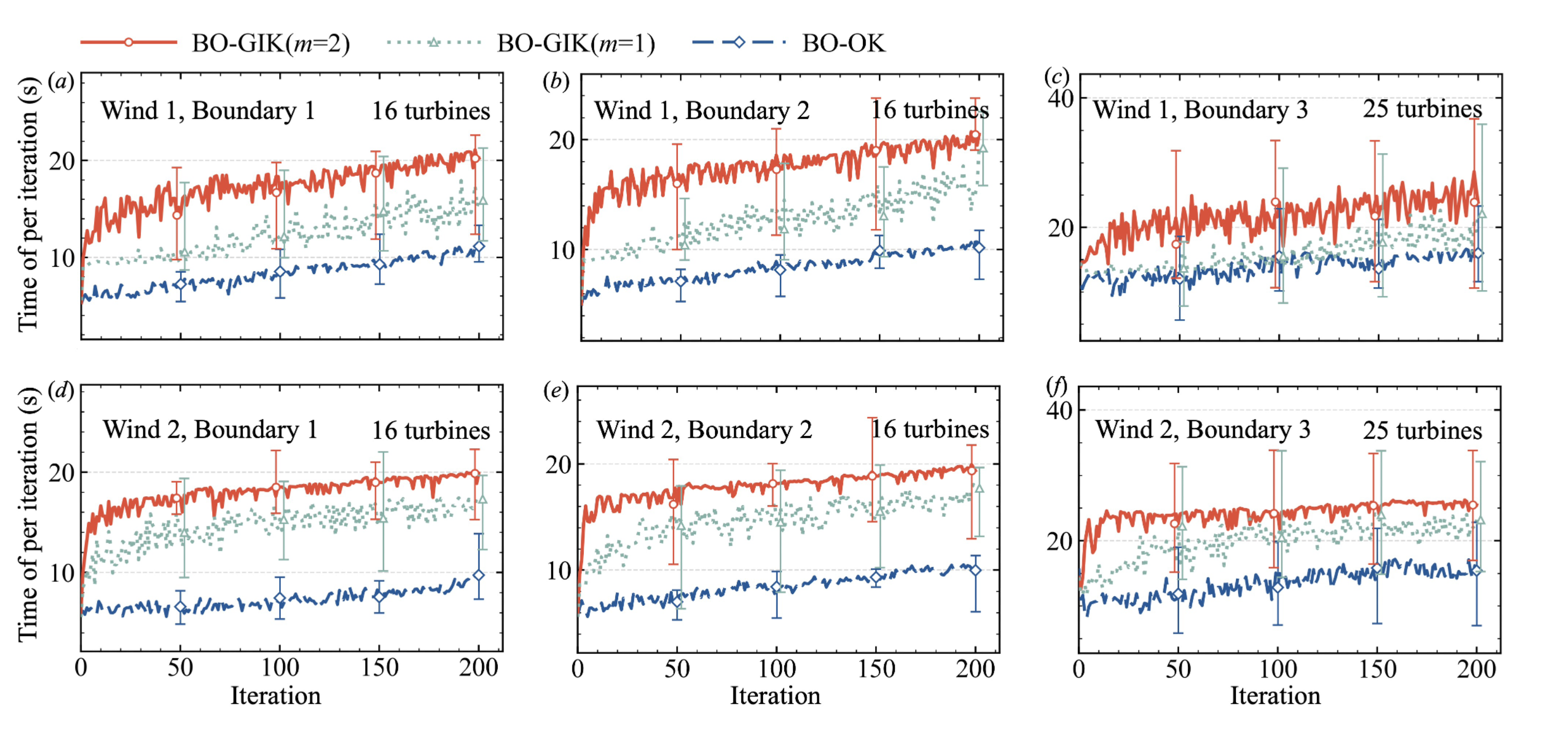}
    \caption{Computational time profiles of BO-GIK and BO-OK across the six optimization cases. ($a$)--($c$) show the results for Wind 1 with Boundaries 1–3, respectively, while ($d$)--($f$) show those for Wind 2 with Boundaries 1–3, respectively. The lines represent the mean computational time per optimization iteration over 10 independent runs, and the bars shown at 50-iteration intervals indicate the corresponding minimum-to-maximum range across the 10 runs.}
    \label{fig:IterationTime}
\end{figure}

The BO-GIK framework consistently exhibits better optimization performance than the conventional BO-OK baseline across all six cases, as shown in Figure \ref{fig:Iteration}. Both BO-GIK variants (\(m=1\) and \(m=2\)) achieve rapid improvements in AEP during the early stages of optimization, whereas BO-OK generally requires substantially more iterations before yielding noticeable gains. Under the same budget of 200 iterations, both BO-GIK variants converge to higher-AEP layouts across all combinations of wind conditions and farm boundaries. BO-GIK with \(m=2\) generally achieves slightly better final solutions than that with \(m=1\), although the difference is modest. This result indicates that, while the higher-order graph terms considerably improve the predictive accuracy of the surrogate, their additional contribution to the final optimization outcome is comparatively limited under the present test conditions. A plausible explanation is that the first-order Graph-informed mean already captures the dominant relative structure of the AEP landscape sufficiently well to guide the acquisition function toward promising regions. The higher-order terms further improve the quantitative representation of the AEP landscape, but these corrections do not necessarily alter the ranking of high-performing candidate layouts to the same extent. This observation suggests that highly accurate global regression may not be a prerequisite for effective Bayesian optimization. Instead, capturing the relative structure and ranking of promising regions can be sufficient to provide informative guidance for sequential sampling, which may explain why even the first-order graph prior substantially improves the sample efficiency of BO.

The computational costs of the BO-GIK and BO-OK frameworks are compared in Figure \ref{fig:IterationTime}. 
As the optimization proceeds, the per-iteration execution time increases for all methods because the growing training dataset increases the computational cost of Gaussian-process training and updating. 
BO-GIK incurs a modest additional overhead relative to BO-OK due to the evaluation of the Graph-informed mean function at each iteration, with the \(m=2\) variant being slightly more expensive than \(m=1\) because of the additional higher-order graph operations. 
Nevertheless, the computational costs of BO-GIK and BO-OK remain of the same order of magnitude throughout the optimization. 
Importantly, the advantage of BO-GIK does not arise from a lower computational cost per BO iteration, but from its improved sample efficiency, as demonstrated in Figure \ref{fig:Iteration}. 
The additional surrogate-modeling overhead is therefore modest relative to the substantial improvement in optimization performance obtained under the same objective-evaluation budget.

Since BO-OK has a lower computational cost per iteration, we further extend the BO-OK optimization to 500 iterations to examine whether the additional iteration budget can compensate for its lower sample efficiency. 
Table \ref{tab:bo_extended_comparison} summarizes the mean optimized AEP and cumulative computation time obtained by BO-GIK after 200 iterations and by BO-OK after 500 iterations. Despite using 2.5 times as many optimization iterations, BO-OK still yields lower mean optimized AEPs than both BO-GIK variants across all six cases. 
Moreover, the cumulative computation time of the 500-iteration BO-OK runs exceeds that of the 200-iteration BO-GIK runs in every case. These results further demonstrate that the advantage of BO-GIK arises from its improved sample efficiency rather than a lower computational cost per iteration: the Graph-informed surrogate enables the optimization to identify higher-performing layouts with substantially fewer sequential evaluations.
\begin{table}[t]
    \centering
    \caption{Optimization performance and computational cost of BO-GIK and the extended BO-OK runs across the six analytical-wake-model WFLO cases. BO-GIK results are reported after 200 iterations, whereas BO-OK results are reported after 500 iterations. The optimized AEP and cumulative computation time are averaged over 10 independent runs.}
    \label{tab:bo_extended_comparison}
    \small
    \setlength{\tabcolsep}{4.5pt}
    \renewcommand{\arraystretch}{1.15}

    \begin{tabular}{cccccccc}
        \toprule
        & &
        \multicolumn{2}{c}{BO-GIK ($m=2$, 200 iter.)}
        & \multicolumn{2}{c}{BO-GIK ($m=1$, 200 iter.)}
        & \multicolumn{2}{c}{BO-OK (500 iter.)} \\

        \cmidrule(lr){3-4}
        \cmidrule(lr){5-6}
        \cmidrule(lr){7-8}

        Wind
        & Boundary
        & AEP (GWh) & Time (min)
        & AEP (GWh) & Time (min)
        & AEP (GWh) & Time (min) \\
        \midrule

        \multirow{3}{*}{1}
        & 1
        & 466.5 & 52.3
        & 465.7 & 38.5
        & 436.4 & 127.1 \\

        & 2
        & 458.0 & 52.9
        & 457.7 & 42.5
        & 426.6 & 128.4 \\

        & 3
        & 729.8 & 65.4
        & 729.2 & 52.9
        & 695.9 & 181.3 \\

        \midrule

        \multirow{3}{*}{2}
        & 1
        & 365.5 & 54.3
        & 365.0 & 49.3
        & 354.7 & 117.9 \\

        & 2
        & 356.7 & 59.8
        & 356.4 & 49.1
        & 345.8 & 128.1 \\

        & 3
        & 576.0 & 81.1
        & 575.0 & 66.7
        & 565.6 & 175.5 \\

        \bottomrule
    \end{tabular}
\end{table}

\begin{figure}
    \centering
    \includegraphics[width=1.0 \textwidth]{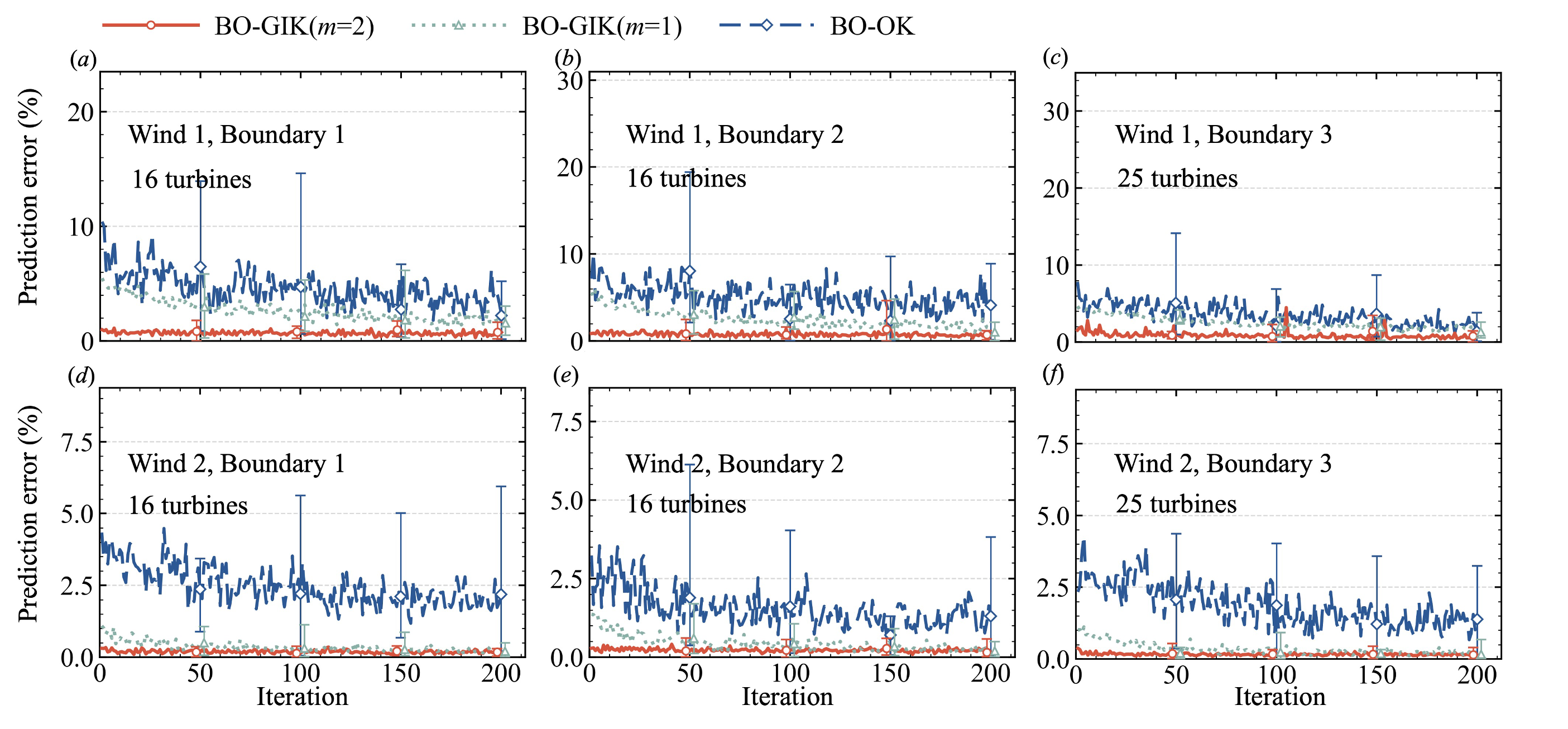}
    \caption{Prediction errors of selected candidate samples during the optimization process. The prediction error is defined as $\left|\widehat{\mathrm{AEP}}-\mathrm{AEP}\right|/\mathrm{AEP}\times100\%$, where $\widehat{\mathrm{AEP}}$ is the surrogate-predicted AEP of the selected candidate before objective evaluation, and $\mathrm{AEP}$ is the corresponding actual value obtained from the wake model.}
    \label{fig:IterationError}
\end{figure}

To further investigate the difference in optimization behavior between BO-GIK and BO-OK, we track the prediction errors at the candidate points selected during the sequential optimization process, as shown in Figure \ref{fig:IterationError}. Across all six cases, BO-OK exhibits relatively large and highly variable prediction errors, although the average error gradually decreases as additional training samples are accumulated. Such prediction inaccuracies reduce the reliability of the surrogate in assessing newly selected candidates, thereby limiting the ability of the acquisition function to consistently identify promising regions of the design space. This behavior helps explain the relatively slow improvement in AEP observed for BO-OK. 
In contrast, BO-GIK (\(m=2\)) maintains substantially lower and more stable prediction errors from the early stages of optimization. The more accurate surrogate predictions provide more reliable guidance for candidate selection, allowing the acquisition function to concentrate evaluations more effectively in high-performing regions and thereby improving the sample efficiency of the optimization.

\subsection{Comparison with established direct optimization methods}
\label{sec:CompareDirect}

\begin{figure}
    \centering
    \includegraphics[width=0.8\linewidth]{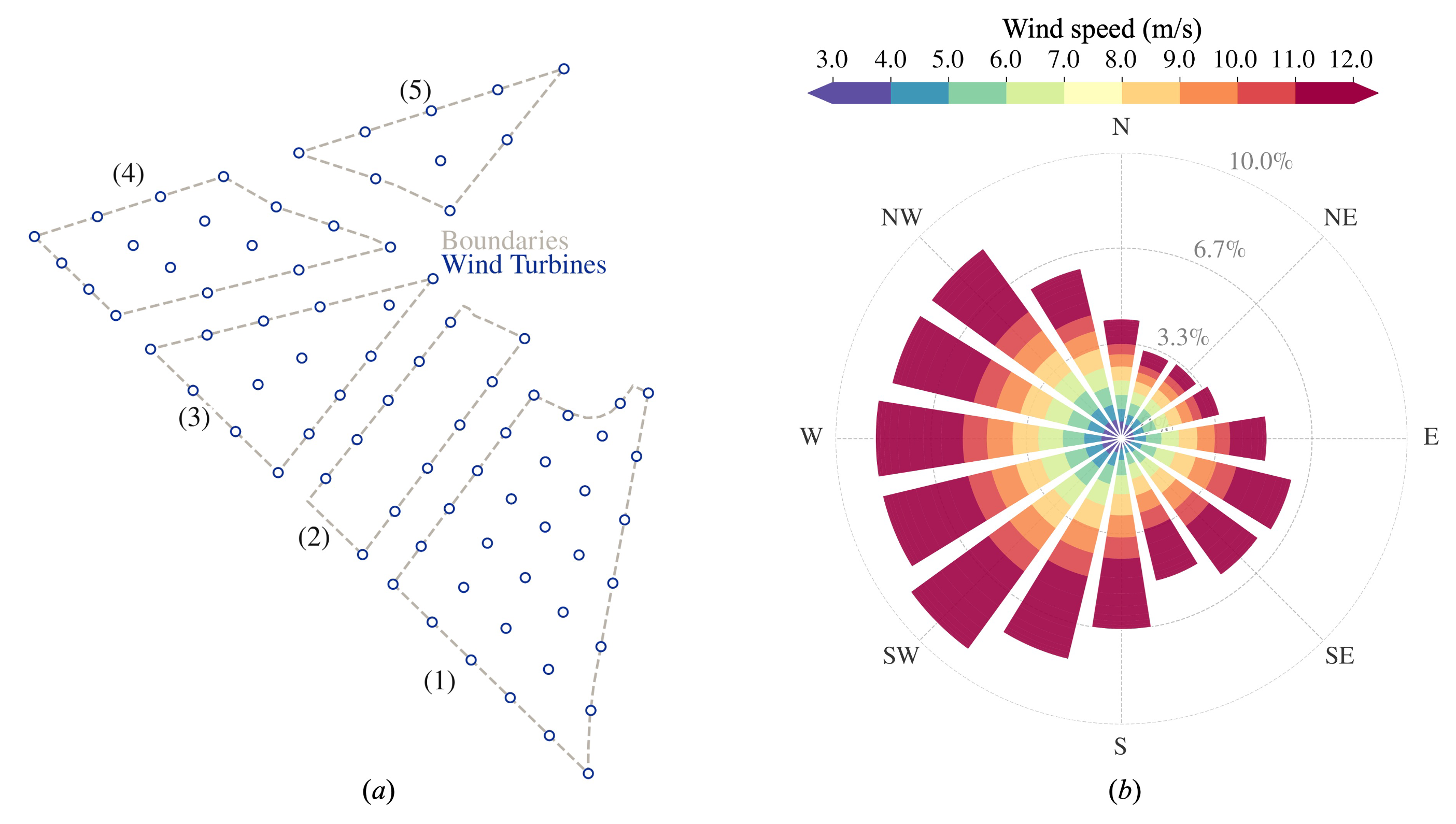}
    \caption{The ($a$) wind farm boundary and ($b$) wind rose for the IEA Wind Task 37 considered by \citet{thomas2023comparison}.}
    \label{fig:81WTsettings}
\end{figure}

Having demonstrated the advantage of BO-GIK over ordinary-Kriging-based Bayesian optimization, we next assess its competitiveness against established direct optimization methods using the IEA Wind Task 37 benchmark studied in \citet{thomas2023comparison}. The benchmark considers the placement of 81 turbines within five disconnected regions, several of which contain concave boundaries, under a multidirectional wind resource, as shown in Figure \ref{fig:81WTsettings}. The resulting problem involves 162 positional design variables and a highly non-convex, multimodal search space. 
\citet{thomas2023comparison} evaluated eight optimization methods using a common objective-function formulation, with each method implemented or guided by researchers experienced with the corresponding algorithm. The benchmark therefore provides a controlled and well-established basis for assessing the performance of BO-GIK against representative direct optimization approaches.

Three representative methods are selected from the eight algorithms considered by \citet{thomas2023comparison}. SNOPT+WEC represents the gradient-based class and was the only strictly gradient-based method considered in the benchmark, achieving competitive AEP with relatively few function calls. DEBO is a discrete optimization method developed specifically for this case study and produced the highest-AEP layout among all eight methods. CMA-ES is selected as a representative general-purpose, gradient-free stochastic optimizer implemented using the publicly available pyCMA library. Together, these three methods provide representative comparisons with gradient-based, problem-specific discrete, and general-purpose evolutionary optimization strategies.

To ensure a consistent objective-function evaluation, BO-GIK (with $m=2$) uses the publicly available physics model and benchmark data released by \citet{thomas2023comparison}. The same Gaussian wake model, wind-resource specification, turbine model, and associated parameters are retained in the present calculations. Thus, the AEP values obtained by BO-GIK are evaluated under the same physical and numerical assumptions as the reference solutions, avoiding discrepancies caused by differences in wake-model implementation or objective-function definition.

The optimization results and computational requirements are summarized in Table \ref{tab:81WTcompare}. The results for SNOPT+WEC, DEBO, and CMA-ES are taken directly from \citet{thomas2023comparison}. SNOPT+WEC was evaluated using 10 different starting layouts, yielding a mean AEP of 2906.3 GWh and a range of 2898.8--2910.1 GWh. CMA-ES was likewise evaluated over 10 independent runs, whereas DEBO was evaluated once because relatively little randomness is involved in the algorithm. To provide a comparable assessment of initialization sensitivity, BO-GIK is also evaluated over 10 independent runs. It should be emphasized that the reported wall-clock times cannot be directly compared because the methods were implemented using different programming environments, hardware platforms, and degrees of parallelization. The BO-GIK calculations are performed on a MacBook equipped with an Apple M2 processor; the corresponding runtime is therefore reported only to indicate the practical computational overhead of the proposed framework rather than to establish a direct speed comparison with the reference methods.

BO-GIK achieves a mean AEP of 2908.6 GWh over 10 independent runs, with optimized solutions ranging from 2907.0 to 2911.2 GWh. The mean BO-GIK result exceeds that of SNOPT+WEC (2906.3 GWh), and even its worst run remains above the best CMA-ES solution (2906.6 GWh). DEBO attains the highest AEP among the methods considered, reaching 2913.2 GWh, but this is only approximately 0.07\% higher than the best BO-GIK solution. The narrow spread across the 10 BO-GIK runs further indicates limited sensitivity to initialization. These results show that BO-GIK achieves solution quality comparable to the best direct optimization methods considered in this benchmark.

More importantly, BO-GIK reaches this level of performance with a substantially smaller objective-evaluation budget. It requires only about 200 additional AEP evaluations per run, compared with 17,626 function calls for the best SNOPT+WEC run, approximately 95,000 for CMA-ES, and 106,065 for DEBO. This corresponds to a reduction of roughly two to three orders of magnitude in objective evaluations. Such sample efficiency is particularly important when each evaluation involves an expensive aerodynamic model, for which optimization methods requiring tens of thousands of evaluations become impractical. The benchmark therefore demonstrates that the Graph-informed surrogate can retain competitive optimization performance under a markedly reduced evaluation budget, providing the basis for its subsequent application to CFD-based WFLO.

\begin{table}[t]
\centering
\caption{Optimization performance and computational cost of GIK and selected direct optimization methods. Results for SNOPT + WEC, DEBO and CMA-ES are taken from \citet{thomas2023comparison}. The '-' indicates that the corresponding quantity is not reported in the original study. Runtime denotes the average time per optimization run where multiple runs are performed. For BO-GIK, the total number of AEP evaluations is decomposed into initial evaluations (81 evaluations) and subsequent infilling evaluations performed during the optimization iterations.}
\label{tab:81WTcompare}
\begin{tabular}{lccccc}
\toprule
Method & Mean AEP (GWh) & Min. AEP (GWh) & Max. AEP (GWh) & Runtime (min) & AEP evaluations \\
\midrule
SNOPT+WEC & 2906.3 & 2898.8 & 2910.1 & 16.5& 17,626 \\
DEBO & -- & -- & 2913.2 & $\sim$111 & 106,065 \\
CMA-ES & -- & -- & 2906.6 & $\sim$60 & $\sim$95,000\\
$\textbf{BO-GIK}$ & $\boldsymbol{2908.6}$ & $\boldsymbol{2907.0}$ & $\boldsymbol{2911.2}$ & $\boldsymbol{\sim60}$ & $\boldsymbol{81+200}$ \\
\bottomrule
\end{tabular}
\end{table}

\subsection{Extension to CFD-based WFLO}
\label{sec:wflo_hf}
The preceding analytical-wake-model studies demonstrate that BO-GIK achieves competitive wind farm layouts with a substantially smaller objective-evaluation budget than established direct optimization methods. 
In this section, we investigate whether this sample-efficiency advantage is retained when the objective function is evaluated using CFD, for which each wind farm AEP calculation requires expensive simulations. This assessment is conducted in two stages. First, a controlled 16-turbine test is conducted to compare BO-GIK (with $m=2$) with both ordinary single-fidelity and multi-fidelity Kriging-based Bayesian optimization (BO-OK and BO-MFK) frameworks which are introduced in detail in Appendix \ref{sec:BO}.  
Direct optimization methods are excluded from this comparison because they typically require $10^4$ -- $10^5$ objective evaluations as shown in the previous section, making direct CFD-based application impractical. 
Second, the proposed BO-GIK framework is applied to a 48-turbine wind farm to demonstrate its scalability and practical applicability for a substantially higher-dimensional CFD-based WFLO problem.
Limited by our computational resources, we have used the ADM-RANS as the higher-fidelity model for the current demonstration. The BO-GIK framework can also be configured to integrate with even more expensive CFD models such as the LES with turbines being simplified as Actuator Line Models (ALM-LES) as adopted in \citet{bempedelis2024data} and \citet{janeippel2026physics}.
The CFD setup for the current ADM-RANS is detailed in Appendix \ref{sec:adm}.

\subsubsection{Benchmark GIK against ordinary and multi-fidelity Kriging model}
To isolate the contribution of the Graph-informed mean function, BO-GIK is compared with ordinary Kriging-based BO (BO-OK) and multi-fidelity Kriging-based BO (BO-MFK) using a 16-turbine CFD benchmark. 
The farm adopts Boundary 1 and Wind 1 defined in Figure \ref{fig:LFsettings}. 
The wind resource is discretized into eight wind states, such that each full-farm AEP evaluation requires eight independent ADM-RANS simulations. 
BO-GIK, BO-OK, and BO-MFK are initialized with the same 16 CFD-evaluated layouts and subsequently run for 100 infill iterations. 
For BO-MFK, the influence of fidelity allocation is investigated by considering three low-to-high fidelity sample ratios, $Q=10$, $20$, and $40$. 

BO-GIK additionally requires an offline pairwise aerodynamic database to construct the Graph-informed mean function. For the present CFD-based cases, this prior information is obtained directly from two-turbine ADM-RANS simulations. 
The relative turbine separation and azimuthal angle are discretized using increments of $1D$ and $10^\circ$, respectively, with interactions beyond $15D$ neglected. Exploiting geometric symmetry results in 270 independent two-turbine configurations for each wind condition. With eight wind states, the complete database therefore consists of $8\times270=2160$ two-turbine ADM-RANS simulations.
These simulations provide the offline training data for the pairwise-interaction surrogate (equation \ref{equ:nestedSurrogate}) used to construct the Graph-informed mean function, as discussed in Section \ref{sec:gikFormulation}.
The database requires approximately 12 hours to construct on a single 64-core workstation equipped with EPYC AMD processors. 
It is important to note that this is a one-time offline cost: once generated, the pairwise database is reused throughout the entire optimization process.
This construction effectively transfers aerodynamic information learned from a low-dimensional pairwise configuration space to the much higher-dimensional full-farm design space through the graph representation.

\begin{figure}
    \centering
    \includegraphics[width=1.0\linewidth]{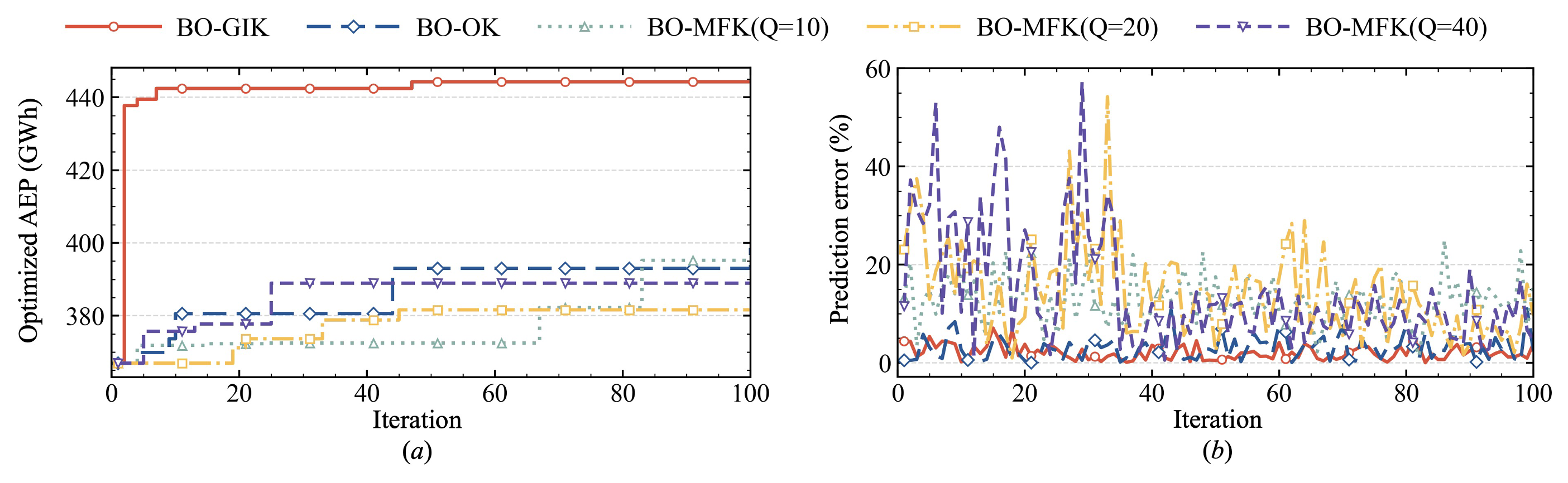}
    \caption{
    Optimization performance comparison among BO-GIK, BO-OK, and BO-MFK for the 16-turbine ADM-RANS-based WFLO case. 
    ($a$) Evolution of optimized AEP during the optimization process. 
    ($b$) Prediction errors of the surrogate models at the selected candidate layouts during sequential optimization. The prediction error is defined as $\left|\widehat{\mathrm{AEP}}-\mathrm{AEP}\right|/\mathrm{AEP}\times100\%$, where $\widehat{\mathrm{AEP}}$ is the surrogate-predicted AEP of the selected candidate before objective evaluation, and $\mathrm{AEP}$ is the corresponding actual value obtained from ADM-RANS.
    BO-MFK results are obtained with different low-to-high fidelity sample ratios ($Q=10$, $20$, and $40$). 
    }
\label{fig:CFD16_compare}
\end{figure}

Figure \ref{fig:CFD16_compare} compares the optimization performance of BO-GIK, BO-OK, and BO-MFK for the 16-turbine ADM-RANS-based WFLO case. 
As shown in Figure \ref{fig:CFD16_compare}($a$), BO-GIK achieves the fastest improvement and consistently obtains higher optimized AEP than BO-OK and BO-MFK under all tested fidelity ratios ($Q=10$, $20$, and $40$). 
The prediction errors in Figure \ref{fig:CFD16_compare}(b) further show that BO-GIK maintains more accurate and stable surrogate predictions during the sequential optimization process, whereas BO-MFK exhibits larger fluctuations despite incorporating additional low-fidelity information. 
These results demonstrate that the developed Graph-informed surrogate provides more effective guidance for CFD-based WFLO than both ordinary Kriging and multi-fidelity Kriging approaches. 

\begin{figure}
\centering
\includegraphics[width=0.85\linewidth]{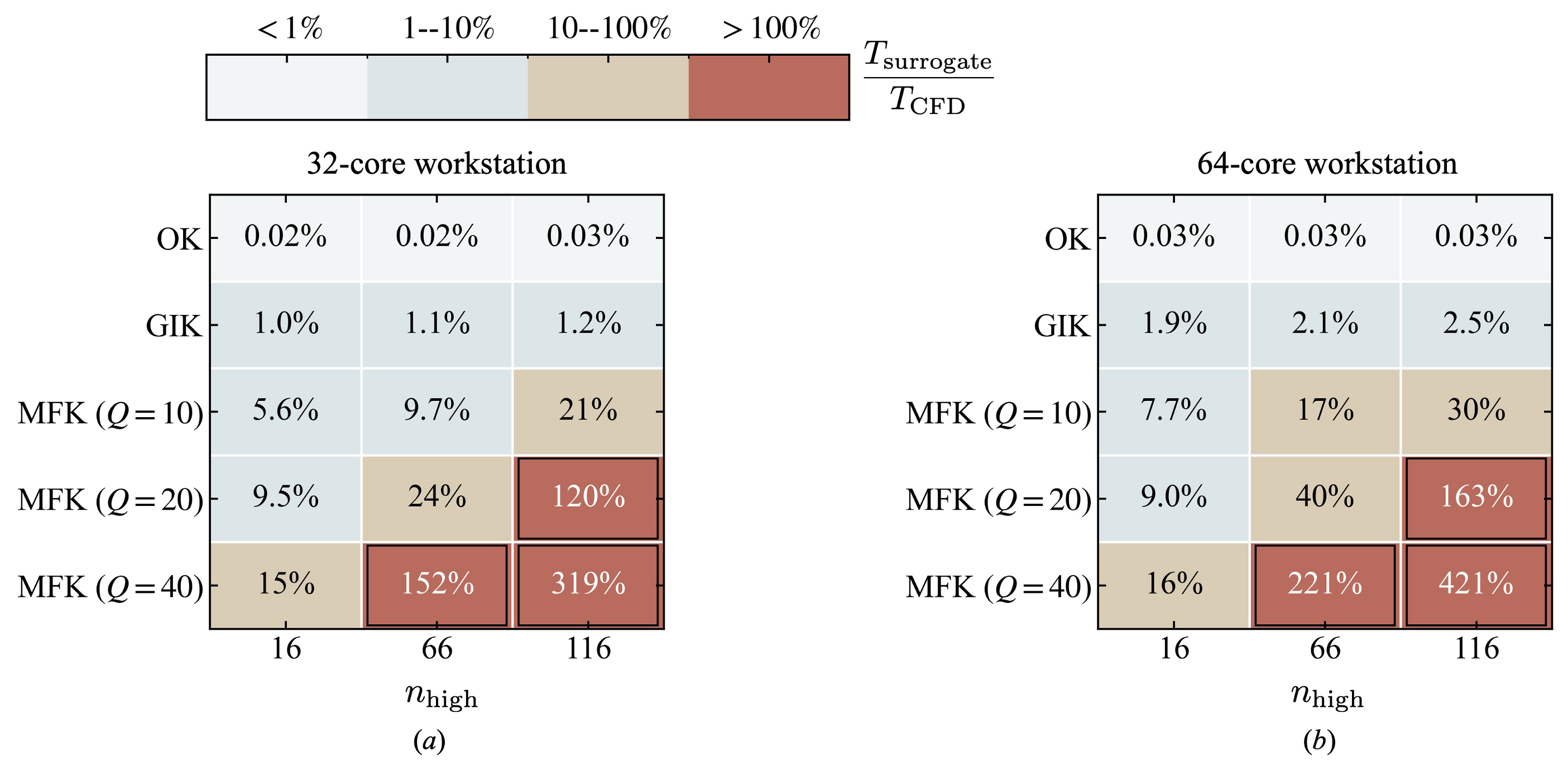}
\caption{Estimated computational cost of surrogate modeling and candidate search in BO, relative to one CFD-based AEP evaluation, for the present implementations of OK, GIK and MFK, with $n_{\mathrm{HF}}=16$, 66, and 116 on the ($a$) 32- and ($b$) 64-core workstations.}
\label{fig:Surrogate_vs_CFD}
\end{figure}

Direct wall-clock time comparisons are not made, as the simulations are performed on different computing platforms (32-core and 64-core workstations). Nevertheless, the MFK method incurs a substantial additional surrogate cost in each BO iteration, mainly arising from the surrogate updates and predictions required to identify promising wind farm layouts. To quantify this overhead, we estimate the BO-related surrogate time as
\begin{equation}
T_{\mathrm{surrogate}} = T_{\mathrm{training}} + n_{\mathrm{pred}} T_{\mathrm{pred}},
\label{eq:surrogate_time}
\end{equation}
with $n_{\mathrm{pred}} = 5000$, consistent with the prediction budget adopted in the optimization. For reference, a single CFD-based AEP evaluation takes on average $T_{\mathrm{CFD}} = 31.4$ min and $18.2$ min on the 32- and 64-core workstations, respectively.
As shown in Figure~\ref{fig:Surrogate_vs_CFD}, the additional cost of OK and the developed GIK remains small over the range of high-fidelity training-set sizes $n_{\mathrm{HF}}$ considered. In contrast, the MFK cost increases sharply with both $n_{\mathrm{HF}}$ and the fidelity ratio $Q$, and can eventually reach or exceed the cost of one complete CFD-based AEP evaluation. 
This significant growth can be explained by the scaling of dense Gaussian-process surrogates. Since the low-fidelity set is enlarged as $n_{\mathrm{LF}}=Q n_{\mathrm{HF}}$, the effective training size of MFK becomes $n_{\mathrm{eff}}=n_{\mathrm{HF}}+n_{\mathrm{LF}}=(1+Q)n_{\mathrm{HF}}$. For a dense GP, training and prediction costs scale as $O(n_{\mathrm{eff}}^{3})$ and $O(n_{\mathrm{pred}}n_{\mathrm{eff}}^{2})$, respectively, so MFK incurs $O((1+Q)^{3}n_{\mathrm{HF}}^{3})$ and $O(n_{\mathrm{pred}}(1+Q)^{2}n_{\mathrm{HF}}^{2})$. In contrast, OK and GIK operate only on the high-fidelity set, giving costs of $O(n_{\mathrm{HF}}^{3})$ and $O(n_{\mathrm{pred}}n_{\mathrm{HF}}^{2})$. 
This difference directly explains the sharp growth of the MFK curve with both $n_{\mathrm{HF}}$ and $Q$, and the much weaker GIK overhead therefore provides an important computational advantage for CFD-based wind farm layout optimization, in addition to its improved optimization performance.

\subsubsection{Larger-scale CFD-based optimization: a 48-turbine wind farm}

\begin{figure}
    \centering
    \includegraphics[width=0.80 \textwidth]{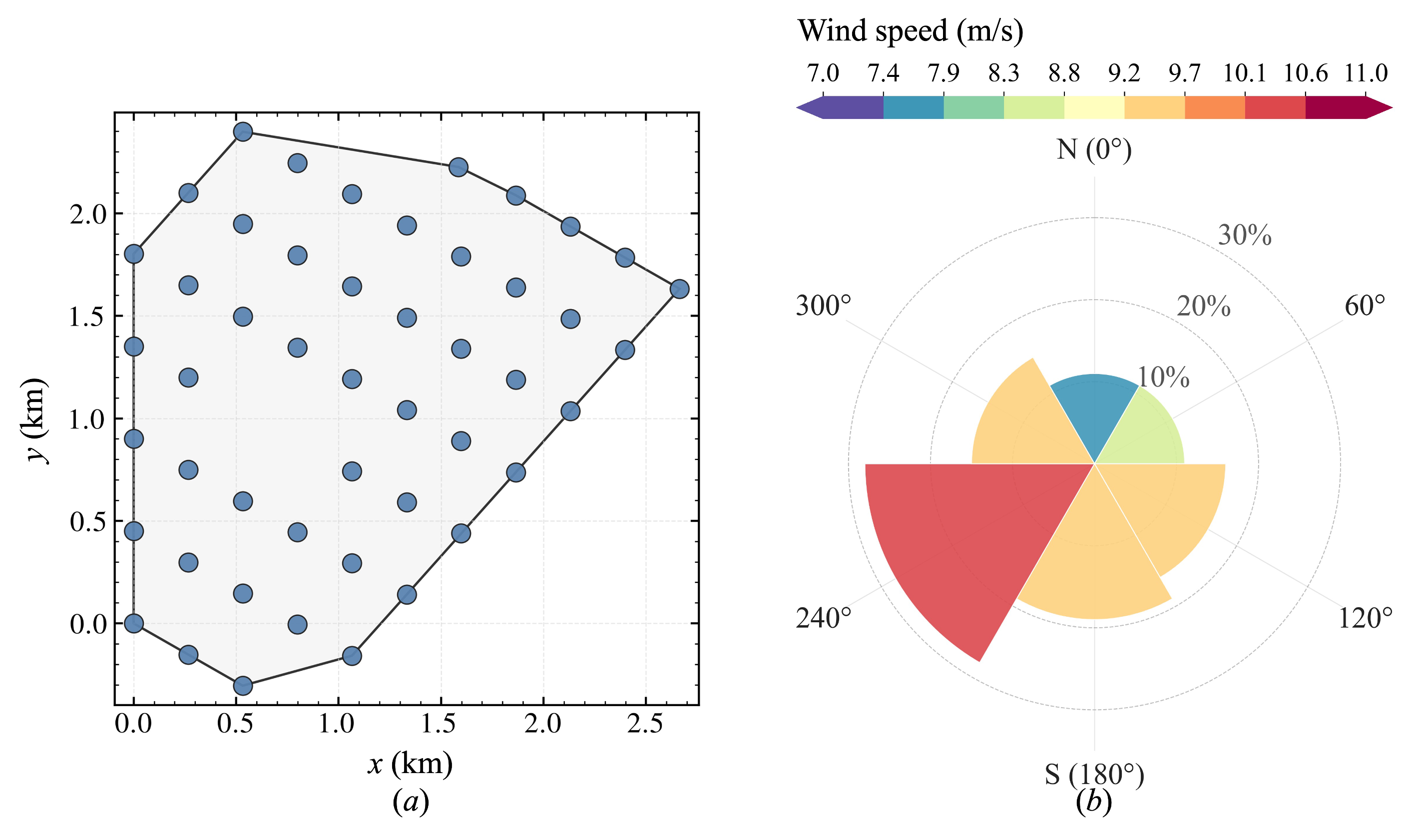}
    \caption{($a$) The wind turbine layout and boundary in Lillgrund wind farm. ($b$) The wind rose for the CFD-based 48-turbine WFLO studies. The turbine positions and original wind data are obtained from \citet{sebastiani2021data}, and the wind data are then grouped into six directional sectors, with the occurrence frequency and mean wind speed of each sector used to define the wind rose.}
    \label{fig:hi_fi_set}
\end{figure}

Having established the advantage of BO-GIK over BO-OK and BO-MFK in the 16-turbine CFD benchmark, we next assess its scalability using a 48-turbine WFLO problem with 96 design variables. To the best of our knowledge, this represents the largest turbine count reported to date for CFD-based WFLO. Rather than repeating the computationally expensive baseline comparisons, this case focuses on whether BO-GIK remains computationally practical at a substantially larger problem scale. As shown in Figure \ref{fig:hi_fi_set}$(a)$, the wind farm boundary is derived from the spatial configuration of the Lillgrund wind farm \citep{pedersen2020integrated,sebastiani2021data}, with the original turbines replaced by NREL 5 MW turbines. The wind resource used for optimization is shown in Figure \ref{fig:hi_fi_set}$(b)$.

All CFD calculations are performed on a single 64-core workstation equipped with AMD EPYC processors, using the numerical setup described in Appendix \ref{sec:adm}. 
The BO-GIK procedure starts from $n_{\mathrm{init}}= 1N = 48$ initial layouts and subsequently performs $n_{\mathrm{infill}}=80$ sequential infill evaluations, giving a total of 128 full-farm AEP evaluations. 
The complete optimization requires approximately two weeks of wall-clock time. 
The optimized layout achieves an AEP of 660.3 GWh, compared with 518.4 GWh for the reference layout in Figure \ref{fig:hi_fi_set}($a$). Since the reference layout uses the larger NREL 5 MW rotor at the original Lillgrund positions, its turbine spacing is compact and its wake losses are severe, so this gain is not intended as a general measure of optimization performance. Rather, this case demonstrates that BO-GIK can handle a 96-dimensional CFD-based WFLO problem with limited full-farm CFD evaluations.

\begin{figure}
    \centering
    \includegraphics[width=1.0\linewidth]{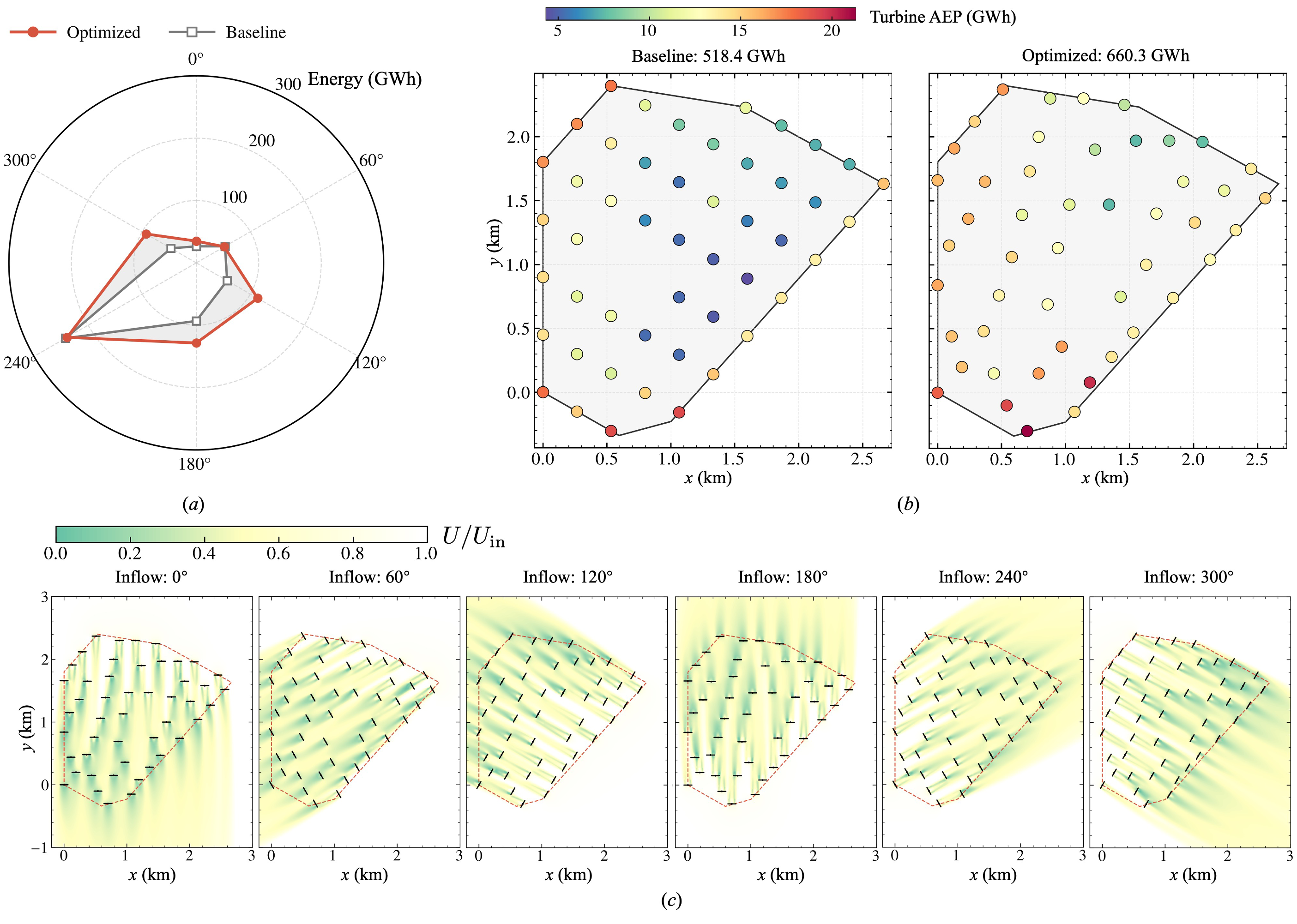}
    \caption{The CFD-based WFLO results obtained using the GIK-based BO framework. ($a$) Directional AEP distributions under different wind conditions. ($b$) Comparison of the baseline and optimized layouts, with turbine-level AEP distributions. ($c$) Velocity contour at the hub height of the optimized layout.}
    \label{fig:HFresult}
\end{figure}

Figure \ref{fig:HFresult} illustrates the resulting redistribution of wind farm performance across the wind rose. The optimized and baseline layouts exhibit comparable energy production under several dominant wind sectors, while larger gains are obtained under other directions. The improvement therefore does not arise from optimization toward a single prevailing inflow condition, but from a redistribution of turbine positions that reduces detrimental wake interactions across multiple wind directions. The turbine-level AEP distributions and hub-height velocity fields further illustrate the corresponding redistribution of energy capture within the farm.

The main computational cost components are summarized in Table \ref{tab:cost_48WT}. As with the 16-turbine case, construction of the offline two-turbine aerodynamic database requires approximately 12 h and represents a one-time preprocessing cost. 
For the full-farm calculations, the wind resource is discretized into six wind states, and one complete CFD-based AEP evaluation requires approximately 2.3 h on the 64-core workstation. 
In contrast, retraining the GIK surrogate using the largest dataset of $n_{\mathrm{train}}=128$ samples requires only approximately 48 s. 
Thus, for the 96-dimensional problem, the overall computational cost remains overwhelmingly dominated by the full-farm CFD evaluations rather than by surrogate construction.

The reported two-week turnaround time corresponds to calculations performed on a single workstation and can be reduced substantially through parallel computing. 
The offline pairwise simulations and the initial full-farm layouts are mutually independent and can be distributed across multiple computing nodes. 
During the sequential BO stage, the six wind-state simulations required for each AEP evaluation are also independent and can be executed concurrently, although the infill iterations themselves remain sequential because each new evaluation is used to update the surrogate before selecting the next candidate. 
With a sufficient number of CPU cores available, the wall-clock cost of each AEP evaluation can therefore be reduced considerably, making the overall optimization substantially faster than the present single-workstation implementation.
{More generally, the GIK framework is not restricted to the present ADM-RANS formulation. The same Graph-informed surrogate strategy can, in principle, be coupled with even higher-fidelity aerodynamic models, such as large-eddy simulation (LES) and actuator-line methods (ALM), provided that the corresponding pairwise-interaction database and full-farm evaluations are available.} 
The improved sample efficiency of GIK is particularly attractive in such settings, where each aerodynamic evaluation is substantially more expensive.

\begin{table}[htbp]
    \centering
    \caption{Computational wall-clock time breakdown for the 48-turbine CFD-based WFLO case.}
    \label{tab:cost_48WT}
    \begin{tabular}{cccc}
        \hline
        \makecell{Two-turbine database\\ preparation} &
        \makecell{Initial CFD\\ evaluations} &
        \makecell{BO infill CFD\\ evaluations} &
        Total \\
        \hline
        $\sim 12$ h &
        $\sim 2.3$ h/eval $\times$ 48 evals & 
        $\sim 2.3$ h/eval $\times$ 80 evals &
        $\sim 306.4$ h \\
        \hline
    \end{tabular}
\end{table}

\section{Conclusions}
\label{sec:conclusion}

This study developed a Graph-informed Kriging (GIK) framework for sample-efficient wind farm layout optimization (WFLO), with particular emphasis on problems involving computationally expensive aerodynamic evaluations. The central idea is to exploit the repeated structure of turbine--turbine aerodynamic interactions rather than treating the full wind farm as an entirely unstructured high-dimensional black-box problem. By representing the wind farm as a directed aerodynamic graph, reusable pairwise interaction information is assembled into a structured mean function for Kriging. The Graph-informed mean captures the dominant farm-scale variation in annual energy production (AEP), while the Gaussian process models the remaining residual structure and uncertainty. This decomposition reduces the amount of global aerodynamic information that must be inferred directly from sparse full-farm training data.

The predictive studies show that GIK consistently provides more accurate AEP estimates than ordinary and generic-trend Kriging models under limited training data. Its advantage is retained as the number of turbines increases and as aerodynamic interactions become stronger. When incorporated into Bayesian optimization, the improved surrogate accuracy translates directly into higher sample efficiency. In the IEA Wind Task 37 benchmark, BO-GIK obtains competitive layouts using only approximately 280 objective evaluations, compared with the much larger evaluation budgets required by the representative direct optimization methods considered in the benchmark. The analytical-wake-model studies further show that the principal advantage of GIK is not a lower computational cost per BO iteration, but the ability to extract more useful optimization information from each expensive objective evaluation.

The benefit of the Graph-informed prior is further demonstrated in CFD-based WFLO. In the controlled 16-turbine ADM-RANS benchmark, BO-GIK provides more accurate and stable surrogate predictions and achieves higher-performing layouts than both ordinary Kriging and multi-fidelity Co-Kriging under the same full-farm high-fidelity evaluation budget. This comparison highlights an important distinction between the approaches: multi-fidelity Kriging relies on additional full-farm low-fidelity observations, whereas GIK transfers reusable low-dimensional pairwise aerodynamic information into the farm-scale surrogate through its structured mean function. The framework is subsequently applied to a substantially larger CFD-based WFLO problem, demonstrating that the proposed approach remains computationally practical as the wind farm scale and design dimensionality increase. To the best of our knowledge, this case represents the largest turbine count reported to date for CFD-based WFLO.

The present framework nevertheless has several limitations. First, the Graph-informed mean is constructed from pairwise turbine interactions and graph-based descriptors of their farm-scale organization. Although this representation captures the dominant AEP variation in the cases considered here, it does not explicitly resolve all multi-turbine aerodynamic effects, and its effectiveness may decrease when strongly non-pairwise flow phenomena dominate the farm response. Second, the present CFD demonstrations employ steady ADM-RANS simulations. Unsteady atmospheric turbulence, wake meandering, turbine-control dynamics, and blade-resolved aerodynamic effects are therefore not represented explicitly. Third, the construction of the Graph-informed mean requires an offline pairwise aerodynamic database. Although this database is reusable throughout optimization and is substantially cheaper than repeated full-farm CFD evaluations, it may need to be regenerated when the turbine model, operating strategy, atmospheric conditions, or aerodynamic fidelity is changed substantially.

Future work will therefore focus on extending the Graph-informed prior to higher-fidelity and time-dependent aerodynamic models, including actuator-line and large-eddy simulations, and on assessing its performance under a broader range of atmospheric and operational conditions. Another important direction is to incorporate additional design objectives and constraints, such as structural loading, control strategies, and economic metrics, into the Graph-informed optimization framework. More broadly, the present results indicate that embedding reusable aerodynamic interaction information into probabilistic surrogates can alleviate the sample-efficiency bottleneck of large-scale CFD-based WFLO, making higher-fidelity aerodynamic models more practical for wind farm design and providing a pathway toward more reliable layouts, reduced wake losses, and improved wind farm energy production.

\section*{Acknowledgments}
The financial supports from the National Key R\&D Program of China (No. 2023YFE0120000), National Natural Science Foundation of China (Nos. 12572267, 52371285), Guangdong Basic and Applied Basic Research Foundation (No. 2023A1515240054) and Program for Intergovernmental International S\&T Cooperation Projects of Shanghai Municipality, China (No. 24510711100) are gratefully acknowledged. 

\appendix
\section{Wind farm modeling}
\subsection{Gaussian wake model}
\label{sec:gaussian} 
In this study, the Gaussian wake model \citep{bastankhah2016experimental,king2021control} is employed as a low-fidelity analytical tool to predict wake behavior. The streamwise velocity in the wake of a turbine, $u_G$, is defined as:
\begin{subequations}
\begin{align}
	& \frac{u_G(x,y,z)}{U_\infty} = 1 - Ce^{-(y-y_0)^2/2\sigma^2_ye^{-(z-z_h)^2/2\sigma_z^2}},\\
	& C = 1-\sqrt{1-\frac{(\sigma_{y0}\sigma_{z0})C_T}{\sigma_y\sigma_z}},
\end{align}
\end{subequations}
where $U_\infty$ is the freestream velocity, $C$ is the wake center's velocity deficit, $\sigma_y$ and $\sigma_z$ represent the wake width in $y$ and $z$ directions, $y_0$ is turbine's lateral position and $z_h$ is the hub height. 
$\sigma_y$ and $\sigma_z$ are computed by
\begin{subequations}
\begin{align}
	& \frac{\sigma_z}{D} = k_z\frac{(x-x_0)}{D}+\frac{\sigma_{z0}}{D},\\
	& \frac{\sigma_y}{D} = k_y\frac{(x-x_0)}{D}+\frac{\sigma_{y0}}{D},
\end{align}
\end{subequations}
where $\frac{\sigma_{z0}}{D}=\frac{\sigma_{y0}}{D} = \frac{1}{2}\sqrt{\frac{u_R}{U_\infty+u_0}}$, $D$ is the rotor diameter, $u_R$ is the velocity at the rotor, $u_0$ is the velocity of far wake, $k_y$ and $k_z$ are the wake expansion coefficients in the lateral and vertical direction.

Once the effective inflow velocity at each turbine is determined from the wake model, the corresponding power coefficient \(C_P\) is obtained from the manufacturer-provided lookup table. The power output of turbine \(i\) is then calculated as
\begin{equation}
P_i = \frac{1}{2}\rho A_D U_{\mathrm{eff},i}^3 C_P(U_{\mathrm{eff},i})e,
\end{equation}
where \(U_{\mathrm{eff},i}\) denotes the effective inflow velocity experienced by turbine \(i\), \(A_D\) is the rotor swept area, \(\rho\) is the air density, and \(e\) is the generator efficiency.

\subsection{ADM-based CFD simulations}
\label{sec:adm}
In this study, we employ the Reynolds-averaged Navier-Stokes (RANS) formulation as the high-fidelity tool to model the wake interactions among wind farms.
Similar to previous CFD-based WFLO studies \citep{cruz2020wind,antonini2020optimal}, the rotors are simplified as actuator disks to reduce the computational complexity. Its governing equations are formulated as follows:
\begin{subequations}
    \label{equ:rans}
    \begin{align}
          \frac{\partial \overline{u}_i}{\partial x_i} & = 0, \\
         \rho\frac{\partial(\overline{u}_i\overline{u}_j)}{\partial x_j} & = -\frac{\partial \overline{p}}{\partial x_i}+\frac{\partial}{\partial x_j}\left(\mu\left(\frac{\partial \overline{u}_i}{\partial x_j}+\frac{\partial \overline{u}_j}{\partial x_i}\right)-\rho\overline{u^{'}_{j} u^{'}_{i}}\right) + f_i,
    \end{align}
\end{subequations}
where $x_{i}$ is Cartesian space coordinate, $\overline{u}_{i}$ and $\overline{p}$ are the temporally averaged velocity and pressure, respectively.
$\rho$ and $\mu$ are the air density and dynamic viscosity, $f_i$ represents the source terms from the actuator disk.
The $k-\epsilon$ turbulence model is then employed to solve the time-average Reynolds stress $\rho\overline{u_{j}^{'}u_{i}^{'}}$.
The standard transport equations for it are as follows:
\begin{subequations}
    \label{equ:turbulence}
    \begin{align}
        \rho\frac{\partial (\overline{u}_{i}k)}{\partial x_i} &=\frac{\partial}{\partial x_j}\left(\frac{\mu_t}{\sigma_k}\frac{\partial k}{\partial x_j}\right)+P_k-\rho\epsilon , \\
        \rho\frac{\partial(\overline{u}_{i}\epsilon)}{\partial x_i} &=\frac{\partial}{\partial x_j}\left(\frac{\mu_t}{\sigma_\epsilon}\frac{\partial\epsilon}{\partial x_j}\right)+C_{1\epsilon}\frac{\epsilon}{k}P_k - C_{2\epsilon}\rho\frac{\epsilon^2}{k} ,
    \end{align}
\end{subequations}
with 
\begin{subequations}
\label{equ:pkmu}
    \begin{align}
        P_k &= -\rho\overline{u_{j}^{'}u_{i}^{'}}\frac{\partial \overline{u}_j}{\partial x_i} , \\
        \mu_t &= \rho C_\mu \frac{k^2}{\epsilon},
    \end{align}
\end{subequations}
where $C_\mu = 0.09$, $\sigma_k=1.0$,  $\sigma_\epsilon=1.3$, $C_{1\epsilon}=1.44$, $C_{2\epsilon}=1.92$ are the five constants in the $k-\epsilon$ turbulence model.

The volumetric source terms  $f_i$ are computed by distributing the thrust and torque following Goldstein optimum \citep{goldstein1929vortex} and the one-dimensional actuator disk model (ADM) \citep{burton2011wind}:
\begin{subequations}
\label{equ:distribution}
    \begin{align}
        f_{ix} &= A_{x}r^*\sqrt{1-r^*}, \\
        f_{i\theta} &= A_\theta \frac{r^*\sqrt{1-r^*}}{r^*(1-r^{'}_{h})+r^{'}_{h}},
    \end{align}
\end{subequations}
with
\begin{subequations}
\label{equ:distributionpara}
    \begin{align}
        r^* &= \frac{r^{'}-r^{'}_h}{1-r^{'}_h}, \ \ r^{'} = \frac{r}{R_P}, \ \ r_h^{'} = \frac{R_H}{R_P}, \\
        A_x &= \frac{105}{8}\frac{T}{\pi\Delta(3R_H+4R_P)(R_P-R_H)}, \\
        A_\theta &= \frac{105}{8}\frac{M}{\pi\Delta R_P(3R_P+4R_H)(R_P-R_H)},
    \end{align}
\end{subequations}
where $f_{ix}$ is axial force, $f_{i\theta}$ is tangential force, $r$ is the distance between the point and disk center, $R_P$ is the external radius of disk, $R_H$ is the internal radius of disk, $T$ is rotor's thrust, $M$ is rotor's torque, and $\Delta$ is the thickness of disk.

In wind farm CFD simulations, wake interactions obscure the effective inflow velocity for downstream turbines, making it difficult to prescribe the correct thrust and torque a priori. To address this challenge, this study adopts an iterative approach to compute the aerodynamic loads. The procedure is detailed as follows:
\begin{subequations}
\label{equ:admtheory}
    \begin{align}
        & U_{D,i} = U_{\mathrm{eff},i}(1-a_i), \\
        & C_{T,i} = \frac{T_i}{\frac{1}{2}\rho A_{D} U_{\mathrm{eff},i}^{2}}, \\
        & C_{T,i} = 4a_i(1-a_i), \\
        & a_i = \frac{1}{2}(1-\sqrt{1-C_{T,i}}),
    \end{align}
\end{subequations}
where $U_{D,i}$ is the disk-averaged velocity of the $i$-th turbine, $U_{\mathrm{eff},i}$ is the corresponding effective inflow velocity upstream of the rotor excluding the turbine's own induction effect, $T_i$ is the rotor thrust, $C_{T,i}$ is the thrust coefficient, $A_D$ is the actuator-disk area, and $a_i$ is the axial induction factor.
Finally, the aerodynamic power extracted by the $i$-th turbine is evaluated from the work rate of the actuator-disk body force over the corresponding disk region:
\begin{equation}
\label{equ:power}
    P_i = -\sum_{c\in\Omega_i}
    \boldsymbol{f}_c \cdot \boldsymbol{u}_c \,\Delta V_c,
\end{equation}
where $\Omega_i$ denotes the set of computational cells within the actuator-disk region of the $i$-th turbine, $\boldsymbol{f}_c$ and $\boldsymbol{u}_c$ are the body-force density and velocity vector in cell $c$, respectively, and $\Delta V_c$ is the cell volume. The negative sign accounts for the fact that the actuator force exerted on the fluid acts opposite to the local flow direction. The solver details are provided in our previous study \citep{wang2024optimization}.

\begin{figure}
    \centering
    \includegraphics[width=1.0 \linewidth]{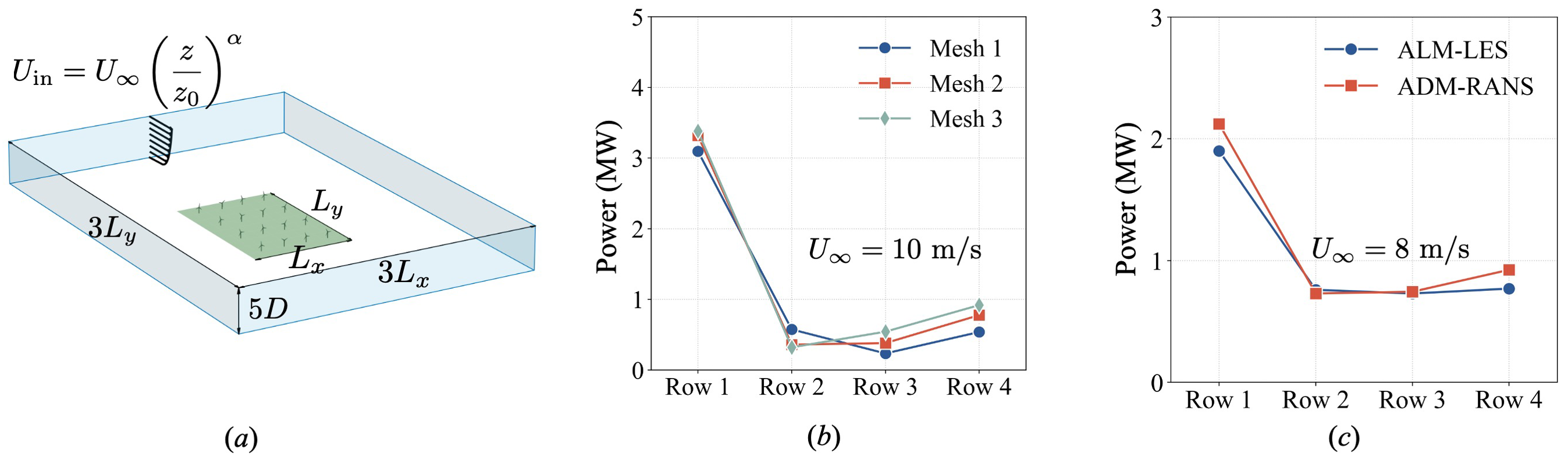}
    \caption{($a$) Setups for wind farm CFD simulations. ($b$) Mesh dependency test. A $4\times 4$ wind farm, characterized by streamwise and spanwise turbine spacings of $5D$ and $3D$, respectively, is used for the test. ($c$) Validation of the predicted row-averaged power output of ADM-RANS against the reference ALM-LES data from \citet{cao2025impact}.}
    \label{fig:hfSettings}
\end{figure}

A rectangular computational domain is employed for the wind farm CFD simulations, which covers $(x, y, z) \in [-1.5L_x, 1.5L_x]\times [-1.5L_y, 1.5L_y]\times [0, 5D]$, where $L_x$ and $L_y$ are determined by the wind farm layouts and the corresponding inflow wind direction, as shown in Figure \ref{fig:hfSettings}($a$).
In the global coordinate system, the $y$-axis aligns with the North ($0^\circ$ inflow) and the $x$-axis with the East.
To simulate varying wind directions, the turbine coordinates are transformed from the global frame $(\boldsymbol{X}, \boldsymbol{Y})$ to a wind-aligned frame $(\boldsymbol{X}', \boldsymbol{Y}')$ for any given inflow angle $\theta_\mathrm{wind}$ (defined as positive clockwise from North). Accordingly, the characteristic lengths $L_x$ and $L_y$ are computed as follows:
\begin{subequations}
\begin{align}
    & x_{\mathrm{max}} = \mathrm{max}\ \boldsymbol{X}' + 2D,\ \  x_{\mathrm{min}} = \mathrm{min}\ \boldsymbol{X}' - 2D, \\
    & y_{\mathrm{max}} = \mathrm{max}\ \boldsymbol{Y}' + 2D,\ \  y_{\mathrm{min}} = \mathrm{min}\ \boldsymbol{Y}' - 2D, \\
    & L_x = x_{\mathrm{max}} - x_{\mathrm{min}},\ \  L_y = y_{\mathrm{max}} - y_{\mathrm{min}}.
\end{align}
\end{subequations}
The inlet velocity is specified as $U_{in} = U_{\infty}(z/z_h)^{\alpha}$, where $z_h = 90$ m is the height of the turbine hub, $U_\infty$ is the reference velocity, and $\alpha = 0.14$ is the shear rate of the velocity profile.
A reference pressure $p_\infty = 0$ with zero gradient condition is set for the outlet velocity.

A mesh dependence study is performed on a $4\times 4$ wind farm, characterized by streamwise and spanwise turbine spacings of $5D$ and $3D$, respectively, and an inflow velocity of 10 m/s is prescribed at the domain inlet.
The minimum element size for the Meshes 1, 2 and 3 are set to be $0.4 D$, $0.15 D$ and $0.1D$ before refinements.
They result in total $2.803\times 10^5$, $8.755\times 10^5$ and $14.713\times 10^5$ control volumes, respectively.
Figure \ref{fig:hfSettings}($b$) shows the average power output of four turbines along rows under different mesh refinement strategies.
The Mesh 1 significantly underpredicts power output. 
While the Meshes 2 and 3 yield similar power output values, the computational cost associated with the Mesh 3 is substantially higher. 
Thus, the resolution for Mesh 2 is adopted for subsequent analysis.

Following the grid independence study, the numerical model is validated against higher-fidelity Large-Eddy Simulation data where rotors are simplified with Actuator Line Models (ALM-LES) from \citet{cao2025impact}. 
The validation case prescribes a neutral atmospheric boundary layer profile with a hub-height wind speed of 8 m/s and an ambient turbulence intensity of $10\%$ to match the wind farm consisting of $4\times4$ turbines in literature. 
To ensure aerodynamic consistency between the two modeling approaches, row-averaged thrust forces from the reference ALM-LES are directly extracted and applied as volumetric force inputs in the present ADM.
Figure \ref{fig:hfSettings}($c$) compares the predicted row-averaged power output with the reference data. 
The ADM-RANS results generally agree well with the LES data.
The model successfully captures the large power drop from Row 1 to Row 2 caused by the wind turbine wake, and the power stabilizes and recovers slightly in Row 3 and 4.
This good agreement shows that the present numerical wind farm CFD simulation framework can be reliably used for the next layout optimization step.

\section{Bayesian optimization and multi-fidelity Kriging}
\label{sec:BO}
\subsection{Bayesian optimization}
\label{sec:BO_AF}
Bayesian optimization (BO) is a sequential optimization strategy for computationally expensive black-box functions \citep{frazier2018tutorial}. 
Starting from an initial set of evaluated designs, BO constructs a probabilistic surrogate of the objective function and iteratively selects additional samples to improve the model and search for better solutions. 
In this study, the Kriging model described in Section~\ref{sec:AEP} is employed as the surrogate. 
For a candidate design $\boldsymbol{z}$, the surrogate provides a posterior mean $f(\boldsymbol{z})$ and a posterior standard deviation $s(\boldsymbol{z})$, representing the predicted objective value and the associated prediction uncertainty, respectively.

The next design to be evaluated is determined by maximizing an acquisition function (AF), which uses the surrogate prediction and uncertainty to guide sequential sampling. 
At the $t$-th BO iteration, the next candidate is selected as
\begin{equation}
    \boldsymbol{z}_{t+1} = \arg\max_{\boldsymbol{z}\in\mathcal{Z}}
    \mathrm{AF}_t(\boldsymbol{z}),
    \label{eq:BO_AF}
\end{equation}
where $\mathcal{Z}$ denotes the feasible layout design space. 
The selected candidate is then evaluated using the original aerodynamic model. 
The new observation is added to the training dataset, the Kriging model is updated, and the procedure is repeated until the prescribed stopping criterion is reached.

Expected Improvement (EI) is a commonly used acquisition function that balances the predicted objective value and surrogate uncertainty \citep{zhan2020expected}. 
For the maximization problem considered here, EI is defined as
\begin{equation}
\mathrm{AF}_{\mathrm{EI}}(\boldsymbol{z})
=
\left[\hat{f}(\boldsymbol{z})-f_{\max}\right]
\Phi\left(
\frac{\hat{f}(\boldsymbol{z})-f_{\max}}
{s(\boldsymbol{z})}
\right)
+
s(\boldsymbol{z})
\phi\left(
\frac{\hat{f}(\boldsymbol{z})-f_{\max}}
{s(\boldsymbol{z})}
\right),
\label{eq:EI}
\end{equation}
where $f_{\max}$ is the best objective value observed so far, while $\Phi(\cdot)$ and $\phi(\cdot)$ denote the cumulative distribution function and probability density function of the standard normal distribution, respectively. 
The first term favors candidates with high predicted improvement, whereas the second term promotes sampling in regions with large surrogate uncertainty.

\subsection{Multi-fidelity Kriging}
\label{sec:BO_MFK}

The multi-fidelity Kriging model incorporates additional low-fidelity aerodynamic simulations  through a classical co-Kriging model \citep{kennedy2000predicting,forrester2007multi}, in which the low- and high-fidelity models are jointly modeled by exploiting their cross-fidelity correlation. 
Specifically, the high-fidelity response is expressed as
\begin{equation}
\hat{f}_H(\boldsymbol{z})=\beta \hat{f}_L(\boldsymbol{z})+\delta(\boldsymbol{z}),
\end{equation}
where $\hat{f}_L(\boldsymbol{z})$ and $\hat{f}_H(\boldsymbol{z})$ denote the low- and high-fidelity responses, respectively, $\beta$ is a scaling coefficient, and $\delta(\boldsymbol{z})$ represents the fidelity discrepancy modeled by a Gaussian process. 
Here, the low-fidelity analytical wake model used in \S \ref{sec:CompareDirect} provides the trend information of the high-fidelity response, while the discrepancy model learns the correction between fidelity levels.

The high-fidelity sample in each BO iteration is selected using the EI criterion in Eq. \ref{eq:EI}.
For the MFK surrogate, the predictive variance $s^2_H(\boldsymbol{z})$ is given by
\begin{equation}
    s^2_H(\boldsymbol{z})=\beta^2s^2_L(\boldsymbol{z})+s^2_\delta(\boldsymbol{z}),
\end{equation}
where $s_L^2(\boldsymbol{z})$ and $s_\delta^2(\boldsymbol{z})$ denote the predictive variances of the low-fidelity and discrepancy models, respectively.
A nested sampling structure is adopted  such that all high-fidelity sample locations are also included in the low-fidelity dataset.
The number of low-fidelity samples is controlled by the prescribed low-to-high-fidelity sample ratio $Q$. 
For each high-fidelity sample, $Q-1$ additional low-fidelity samples are introduced, half of which are generated by local random perturbations around the high-fidelity sample, while the remaining samples are randomly generated over the feasible design space.
This combination of local refinement and global exploration follows common sampling strategies in surrogate-based optimization \citep{regis2013combining,PRENTZAS2026118658}.

\bibliographystyle{unsrtnat}
\bibliography{reference}

\end{document}